\pdfoutput=1
\PassOptionsToPackage{pdfauthor={Christian J. Schuler, Michael Hirsch, Stefan Harmeling, Bernhard Schölkopf},
            pdftitle={Learning to Deblur},
            pdfkeywords={machine learning, blind deconvolution, image processing, neural networks, deep learning},
            colorlinks}{hyperref}
\documentclass[letterpaper,fontsize=12]{scrartcl} 
\usepackage{scrpage2} 
\usepackage[T1]{fontenc} 
\usepackage[utf8]{inputenc} 
\usepackage[title,titletoc,toc]{appendix} 

\usepackage{color} 
\usepackage{amsmath} 
\usepackage{amsfonts} 
\usepackage{amssymb} 
\usepackage{amsthm} 
\usepackage{dsfont} 
\usepackage{textcomp} 
\usepackage{tgtermes} 
\usepackage{helvet} 
\usepackage{}
\usepackage[style=alphabetic,backend=biber,maxcitenames=2,maxbibnames=4,firstinits=true,bibencoding=utf8]{biblatex} 
\usepackage{graphicx} 
\usepackage{caption} 
\usepackage{rotating} 
\usepackage{xspace} 
\usepackage{enumitem} 
\usepackage{multicol} 
\usepackage{tikz} 
\usepackage{xcolor} 
\usepackage{upgreek} 
\usepackage{bm} 
\usepackage{etoolbox} 
\usepackage{todonotes} 
\usepackage{scrhack} 

\graphicspath{{Figures/}{./}}

\DeclareGraphicsExtensions{.jpg, .png, .pdf}   

\pdfoutput=1

\newlength{\singleplotwidth} 
\newlength{\doubleplotwidth} 
\newlength{\singlepw}
\newlength{\doublepw}
\newlength{\triplepw}
\newlength{\quadpw}
\newlength{\quintpw}
\newcommand{\tightcolsep}{\renewcommand\tabcolsep{1.5pt}}

\newlength{\landscapepw}
\makeatletter
\patchcmd{\@maketitle}{\huge}{\LARGE}{}{}
\makeatother
\pdfoutput=1

\newcommand{\vv}[1]{\mathbf{#1}}         
\newcommand{\VV}[1]{\bm{#1}}     

\pdfoutput=1

\DeclareMathOperator{\Diag}{Diag}               

\renewcommand{\H}{\mathsf{H}}                   
\newcommand{\conj}[1]{\overline{#1}}            

\newcommand{\T}{\mathsf{T}}                     

\newcommand{\F}{F}

\pdfoutput=1

\newcommand*{\eg}{e.g.\@\xspace}
\newcommand*{\ie}{i.e.\@\xspace}
\newcommand*{\cf}{cf.\@\xspace}

\makeatletter
\newcommand*{\etc}{%
    \@ifnextchar{.}%
        {etc}%
        {etc.\@\xspace}%
}
\newcommand*{\etal}{%
    \@ifnextchar{.}%
        {et al}%
        {et al.\@\xspace}%
}
\makeatother

\AfterPreamble{\hypersetup{
    linkcolor=[rgb]{0.65,0.035,0.087},
    citecolor=[rgb]{0,0.298,0.651},
    urlcolor=[rgb]{0,0.298,0.651}
}}

\begin{document}

\pdfoutput=1
\title{Learning to Deblur}
\author{\normalsize{
Christian J. Schuler,
Michael Hirsch,
Stefan Harmeling,
Bernhard Schölkopf}\\[-0.5ex]
\normalsize{\texttt{\{cschuler,mhirsch,harmeling,bs\}@tuebingen.mpg.de}}\\
\small{Max Planck Institute for Intelligent Systems}}
\date{}

\maketitle

\begin{abstract}
\vspace*{-4ex}
We describe a learning-based approach to blind image deconvolution. It uses a deep layered architecture, parts of which are borrowed from recent work on neural network learning, and parts of which incorporate computations that are specific to image deconvolution. The system is trained end-to-end on a set of artificially generated training examples, enabling competitive performance in blind deconvolution, both with respect to quality and runtime.
\end{abstract}

\section{Introduction}

Blind image deconvolution is the task of recovering the underlying
sharp image from a recorded image corrupted by blur and noise.
Examples of such distortions are manifold: In photography, long
exposure times might result in camera shake, often in combination with
image noise caused by low light conditions.  Lens aberrations can create blur
across images in particular for wide apertures.  In astronomy and remote sensing, atmospheric
turbulence blurs images.  All these image reconstruction tasks are
examples of \emph{inverse problems}, which are particularly hard to
solve since not only the underlying image is unknown, but also the blur.

We assume that the blurred image $\vv{y}$ is generated by linearly transforming the
underlying image $\vv{x}$ (sometimes called the ``true image'' or ``latent image'')
by a convolution (denoted by $*$), and additive noise $\vv{n}$,
\begin{align}
  \label{eq:bn.1}
  \vv{y} = \vv{k}*\vv{x} + \vv{n}.
\end{align}
The task of blind image deconvolution is to recover $\vv{x}$ given only the blurry image
$\vv{y}$, without knowing $\vv{k}$.  A number of approaches have recently been proposed
to recover the true image from blurred photographs, e.g.,
\cite{Fergus2006,cho2009fast,Xu2010}.  Usually these methods assume some sparsity
inducing image prior for $\vv{x}$, and follow an iterative, multi-scale
estimation scheme, alternating between blur and latent image estimation.

The idea of the proposed method is to ``unroll'' this reconstruction procedure and
pose it as a nonlinear regression problem, where the optimal parameters are
learned from artificially generated data.  As a function approximator, we use a layered architecture akin to a deep neural network or multilayer perceptron. Some of the layers are convolutional, as popular in many recent approaches, while others are non-standard and specific to blind deconvolution. Overall, the system is inspired by \autocite{bottou1991framework} who formulated the idea of neural networks (NNs) as
general processing schemes with large numbers of parameters.

Using extensive training on a large image dataset
in combination with simulated camera shakes, we train the blind
deconvolution NN to
automatically obtain an efficient procedure to approximate the true
underlying image, given only the blurry measurement.

A common problem with NNs, and in particular with large custom built architectures, is that the devil is in the details and it can be nontrivial to get systems to function as desired. We thus put particular emphasis on describing the implementation details, and we make the code publicly available.

\paragraph{Main contributions:} We show how a trainable model can be
designed in order to be able to \emph{learn} blind deconvolution. Results
are comparable to the state of the art of hand-crafted approaches and go beyond
when the model is allowed to specialize to a particular image category.

\section{Related work}
\label{sec:bn.related}

Some recent approaches to blind deconvolution in photography have already been mentioned
above. We refer the interested
reader to~\autocite{wipf2013bayesian} for an overview of the subject, and focus
only on how NNs have been previously used for deconvolution.

Neural networks have been used extensively in image processing.
Comprehensive reviews are \autocite{egmont2002image,de2003nonlinear}, both
of which present broad overviews of applying NNs to all
sorts of image processing operations, including segmentation, edge
detection, pattern recognition, and nonlinear filtering.

Image deconvolution, often also called image reconstruction, has
been approached with NNs for a long time.  However, these
approaches are quite different from our proposed work:
\begin{itemize}
\item \emph{NN to identify the type of blur:} \autocite{aizenberg2006blur} and similarly \autocite{khare2011image}
  apply NNs to identify the type of blur from a restricted
  set of parametrized blurs, \eg~linear-motion, out-of-focus and
  Gaussian blur, and possibly their parameters.

\item \emph{NN to model blurry images:}
  \autocite{cho1991blur} models the blurry image as the result of a neural
  network where at the different layers the blur kernel and the true
  image appear.

\item \emph{NN to inversely filter blurry images:}
  \autocite{tansley1996neural} learns an inverse filter represented by a
  NN to deblur text.

\item \emph{NN to optimize a regularized least squares
    objective:} \autocite{steriti1994blind} proposes also the common
  two-stage procedure to first estimate the blur kernel and then to
  recover the image.  For both tasks Hopfield networks are employed to
  solve the optimization problems.

\item \emph{NN to remove colored noise:}
  The previous chapter presented a method for non-blind deblurring
  that starts with a straight-forward division in Fourier space and
  then removes the resulting artifacts (mainly colored noise) with large neural
  networks.
\end{itemize}
Other learning-based approaches for blind deconvolution
try to learn the deconvolution solution for a single image,
in particular they attempt to learn an appropriate sparse
representation for a given blurry image, \eg \autocite{hu2010single}.  In
our work, we follow a different strategy: instead of learning the
solution or part of a solution for a single fixed image, we use a neural
network to learn a general procedure that is directly applicable to
other images and to different blurs.  Closest to our approach is the
work of \autocite{schmidt2013discriminative}, who train a deblurring
procedure based on regression tree fields.  However, even though their
approach is not limited to a specific blur or a specific image, they consider only the
problem of \emph{non-blind} deblurring, \ie their method assumes that
the blur kernel is known.  This is in contrast to our contribution,
which demonstrates how to train a NN to solve the \emph{blind} deconvolution
problem.  This is a much harder problem, since not only do we have to
solve an underdetermined \emph{linear} problem (originating from the
least-squares formulation of non-blind deconvolution), but
an underdetermined \emph{bilinear} problem, which appears since the
entries of the unknown blur kernel $\vv{k}$ and the pixel values of the unknown
image $\vv{x}$ are multiplied by the convolution, see Eq.\nobreakspace \textup {(\ref {eq:bn.1})}.

Finally we note that \emph{deconvolutional networks}, introduced in
\autocite{zeiler2010deconvolutional} and further extended in
\autocite{zeiler2011adaptive}, are not architectures for image deconvolution. Instead, they use convolutions to link sparse features to
given images.  Their goals are good image representations for tasks such as object recognition and image denoising.

\section{Blind deconvolution as a layered network}

Existing fast blind deconvolution methods work by alternating between the
following three steps \autocite{xu_cvpr2013,Xu2010,cho2009fast}:
\begin{enumerate}
\item \emph{Feature extraction} computes image representations that
  are useful for kernel estimation.  These representations may simply
  be the gradient of the blurry image and the gradient of the current
  estimate, possibly thresholded in flat regions \autocite{Xu2010};
  they may also be preliminary estimates of the sharp image, for
  instance computed by heuristic nonlinear filtering \autocite{cho2009fast}.
\item \emph{Kernel estimation} uses the extracted features to estimate the
  convolution kernel.
\item \emph{Image estimation} attempts to compute an approximation of the sharp
  image using
  the current kernel estimate.
\end{enumerate}
We will represent these steps by a trainable deep neural network, thus adding
more flexibility to them and allowing them to optimally adapt to the
problem.  The layers of the network alternate between (1) a local
convolutional estimation to extract good features, (2) the estimation
of the blur kernel, globally combining the extracted features, and
finally (3) the estimation of the sharp image. Parts (2) and (3) are
fixed (having only one hyper-parameter for regularization).  The free
parameters of the network appear in part (1), the feature
extraction module. Thus, instead of having to learn a model on the
full dimensionality of the input image, which would not be doable using realistic training set sizes, the learning problem is
naturally reduced to learning filters with a limited receptive field.

\subsection{Architecture layout}

\begin{sidewaysfigure}
  \centering
  \includegraphics{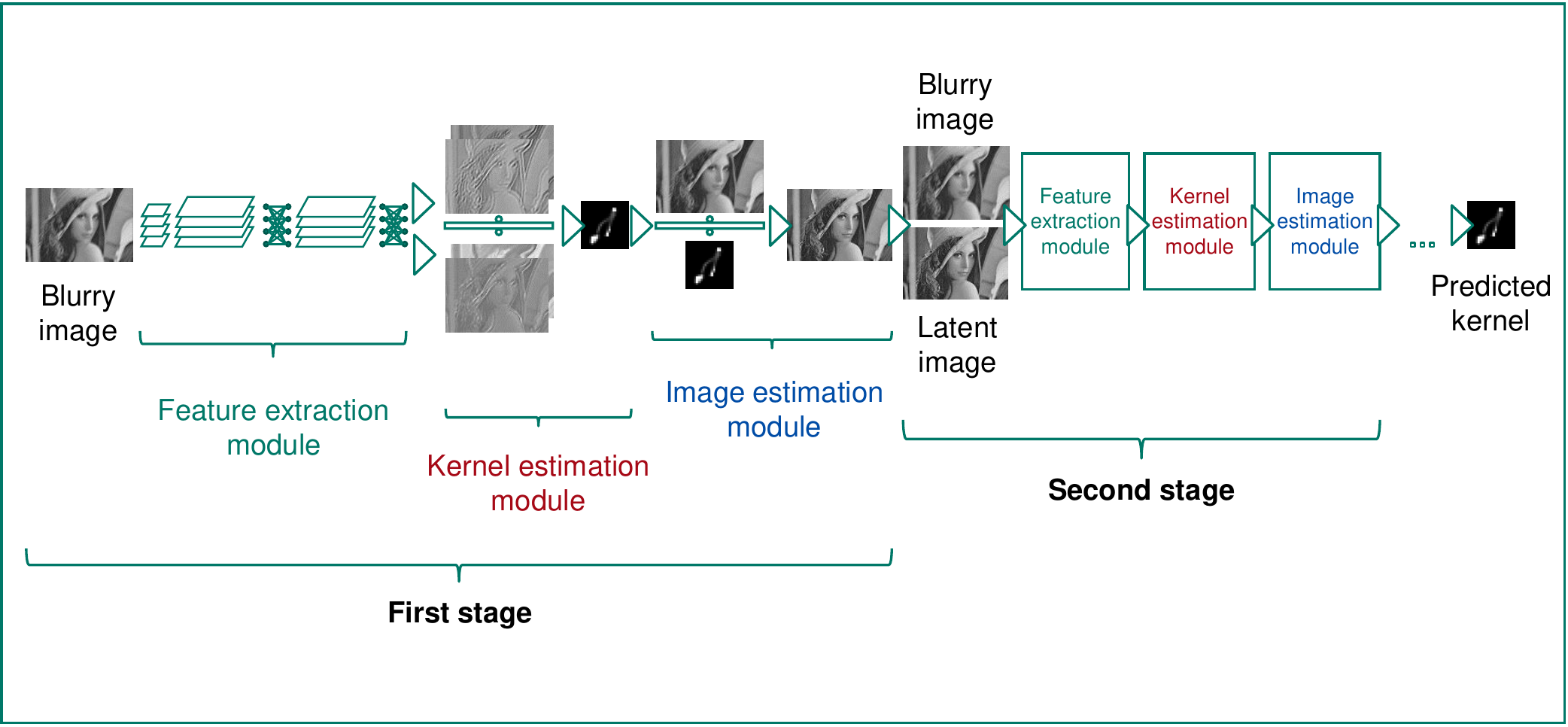}
  \caption[Architecture of our proposed blind deblurring network.]{Architecture of our proposed blind deblurring network. First the \emph{feature extraction module} transforms the image to a learned gradient-like representation suitable for
  kernel estimation. Next, the kernel is estimated by division in Fourier space,
  then similarly the latent image. The next stages, each consisting of these three operations, operate on both the blurry image and the
  latent image.}
\label{fig:illustration}
\end{sidewaysfigure}

\begin{sidewaysfigure}
  \centering
  \includegraphics{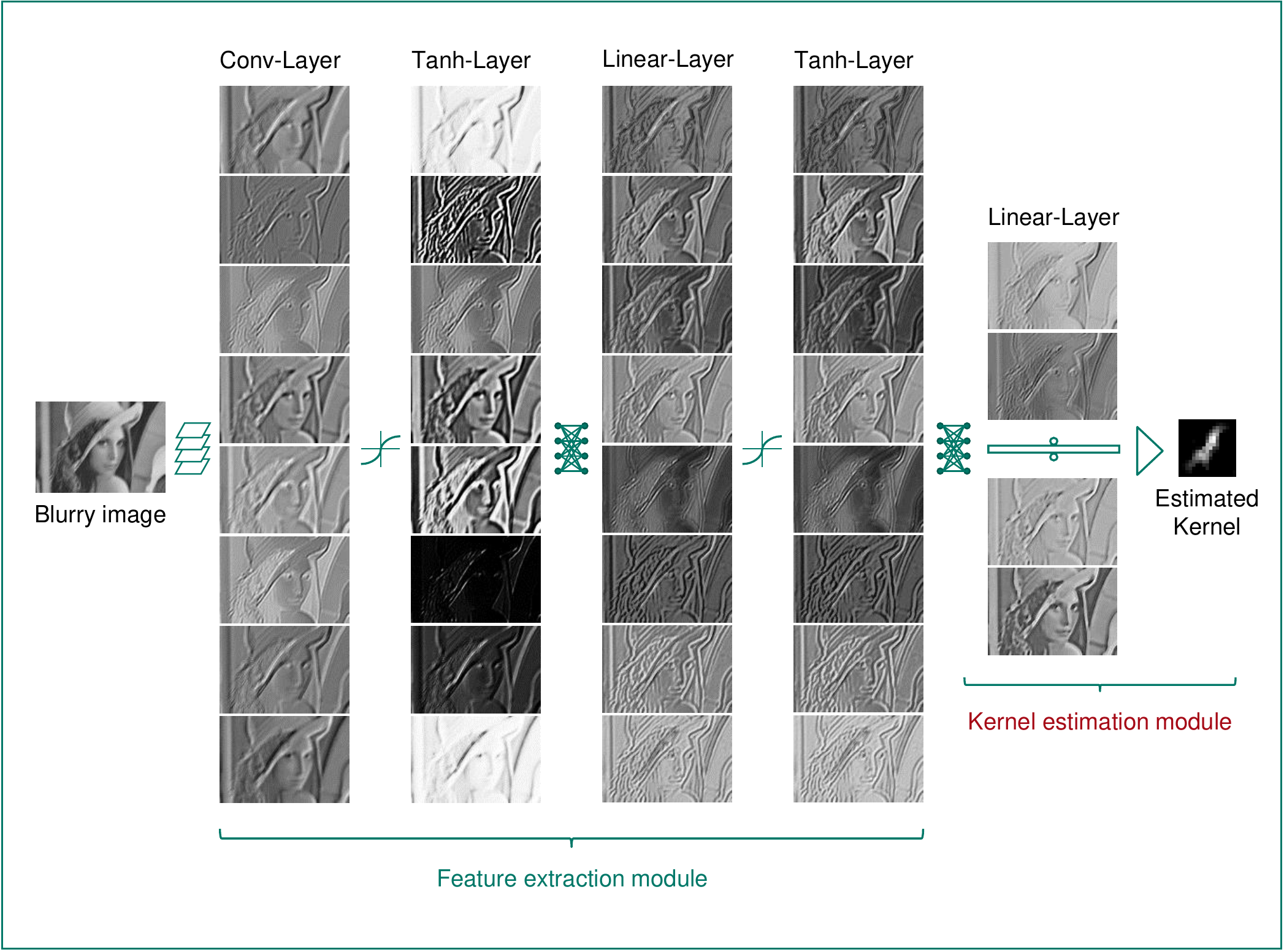}
  \caption[Intermediary outputs of a single-stage NN.]{Intermediary outputs of a single-stage NN with architecture 8$\times$ Conv, Tanh, 8$\times$8, Tanh, 8$\times$4.}
\label{fig:bn.intermediary}
\end{sidewaysfigure}

Below, we describe the different parts of the network.

\subsubsection{Feature extraction module}
What makes a good feature for kernel estimation can reasonably be assumed to be a translation invariant problem, \ie, independent from the position
within the image. We therefore model the feature extractors using shared weights applying across all image locations, \ie, as a convolutional
NN layer,\footnote{Note that this convolution has nothing to do with the convolution appearing in our image formation model (Eq.\nobreakspace \textup {(\ref {eq:bn.1})}) --- causally, it goes in the opposite direction, representing one step in the inverse process turning the blurry image into an estimate of the underlying image.} creating several feature representations of the
image. This is followed by two layers that introduce nonlinearity
into the model. First, every value is transformed by a tanh-unit, then
the feature representations are pixel-wise linearly combined to new
hidden images, formally speaking
\begin{align}
  \label{eq:bn.2}
  \tilde{\vv{y}_i} &= \sum_j \alpha_{ij} \tanh(\vv{f}_j * \vv{y}) & \text{ and }&&
  \tilde{\vv{x}_i} &= \sum_j \beta_{ij} \tanh(\vv{f}_j * \vv{y})
\end{align}
where $\vv{f}_j$ are the filters of the convolution layer (shared for
$\tilde{\vv{x}_i}$ and $\tilde{\vv{y}_i}$), the function $\tanh$ operates
coordinate-wise, and $\alpha_{ij}$ and $\beta_{ij}$ are the
coefficients to linearly combine the hidden representations.  Note
that we usually extract several gradient-like images $\tilde{\vv{x}_i}$ and $\tilde{\vv{y}_i}$.
Depending on the
desired nonlinearity, these two layers can be stacked multiple times,
leading to the final image representations based on features useful
for kernel estimation.

\subsubsection{Kernel estimation module}
Given $\tilde{\vv{x}_i}$ and $\tilde{\vv{y}_i}$ which contain features tuned for optimal
kernel estimation, the kernel $\tilde{\vv{k}}$ can be estimated by
minimizing
\begin{align}
  \sum_i \|\tilde{\vv{k}}\ast \tilde{\vv{x}}_i - \tilde{\vv{y}}_i\|^2 + \beta_k \|\tilde{\vv{k}}\|^2
  \label{eq:bn.k_objective}
\end{align}
for $\tilde{\vv{k}}$ given the results from the previous step $\tilde{\vv{x}_i}$
and their blurry counterparts $\tilde{\vv{y}_i}$. Assuming no noise and a
kernel without zeros in its power spectrum, the true gradients of the
sharp image $\vv{x}$ and its blurred version $\vv{y}$ would return the true
kernel for $\beta_k=0$. Typically, in existing methods $\tilde{\vv{y}}_i$ are just
the gradients of the blurry image, while here these can also be
learned representations predicted from the previous layer. The
minimization problem can be solved in one step in Fourier space if we
assume circular boundary conditions \autocite{cho2009fast}:
\begin{align}
  \tilde{\vv{k}} = F^\H\frac{\sum_i\overline{F \tilde{\vv{x}}_i}\odot F\tilde{\vv{y}}_i}{\sum_i|F\tilde{\vv{x}}_i|^2+\beta_k}.
\label{eq:bn.quotient_layer}
\end{align}
Here $F$ is the discrete Fourier transform matrix, $\odot$ the Hadamard product, $\overline{\vv{v}}$ the complex conjugate of a vector $\vv{v}$, and the division and absolute
value are performed element-wise.
The one step solution is only possible because we use a simple Gaussian prior on the kernel.
We call this step the \emph{quotient layer}, which is an uncommon
computation in NNs that usually only combine linear layers
and nonlinear thresholding units.  The final kernel is returned by
cropping to the particular kernel size and thresholding negative values. To reduce
artifacts from the incorrect assumption of circular boundary
conditions, the borders of the image representations are
weighted with a Barthann window such that they smoothly fade to zero.

Due to varying size of the input image, we set $\beta_k=10^{-4}$ for numerical stability only. This forces the network to not rely on the prior,
which would lose importance for a larger input image relative to the likelihood
term (since the kernel size is fixed).

\subsubsection{Image estimation module}
Before adding another \emph{feature extraction module}, the estimated
kernel is used to obtain an update of the sharp latent image:
analogously to Eq.\nobreakspace \textup {(\ref {eq:bn.k_objective})}, we solve
\begin{align}
  \|\tilde{\vv{k}}\ast \tilde{\vv{x}} - \vv{y}\|^2 + \beta_x \|\tilde{\vv{x}}\|^2
  \label{eq:bn.x_objective}
\end{align}
for $\tilde{\vv{x}}$, which can also be performed with a quotient layer. This can be done in one step (which would not be possible when using a sparse prior on $\tilde{\vv{x}}$).
The following convolution layer then has access to both the latent
image and the blurry image, which are stacked along the third dimension (meaning
that the learned filters are also three-dimensional). The hyper-parameter $\beta_x$
is also learned during training.

\subsection{Iterations as stacked networks}

\begin{figure}[p]
  \centering
  \small
  \includegraphics{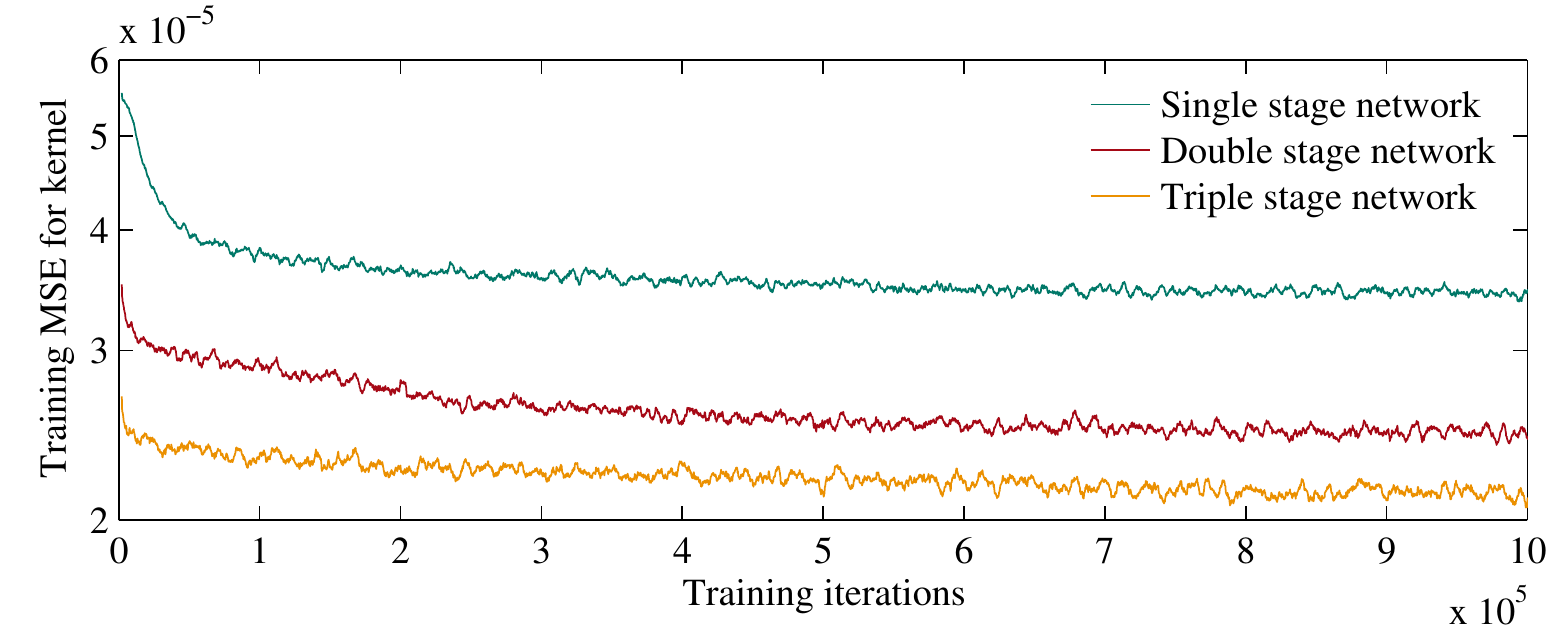}
  \caption[Training curves depending on number of stages.]{Deeper networks are better at kernel prediction. One stage
for kernel prediction consists of a convolutional layer, two hidden layers and a kernel estimation
module.}
\label{fig:bn.depth}
\end{figure}

The \emph{feature extraction module}, \emph{kernel estimation module} and the
\emph{image estimation module} can be stacked several times, corresponding to
multiple iterations in non-learned blind deconvolution methods. This leads to a
single NN that can be trained end-to-end with back-propagation
by taking the derivatives of all steps (see Appendix~\ref{app:math_details} for the
derivatives of the solutions to Eq.\nobreakspace \textup {(\ref {eq:bn.k_objective})} and the analogous
Eq.\nobreakspace \textup {(\ref {eq:bn.x_objective})}), increasing the performance as shown in Fig.\nobreakspace \ref {fig:bn.depth} (but at the same time increasing
runtime).

Similar to other blind deconvolution approaches, we also use a multi-scale
procedure. The first (and coarsest) scale is just a network as described above, trained for a
particular blur kernel size. For the second scale, the previous scale network is
applied to a downsampled version of the blurry image, and its estimated latent
image is upsampled again to the second scale. The second scale network then
takes both the blurry image and the result of the previous scale as input. We
repeat these steps until the final scale can treat the desired blur kernel size
on the full resolution image. 

\begin{figure}[p]
  \centering
  \small
  \includegraphics{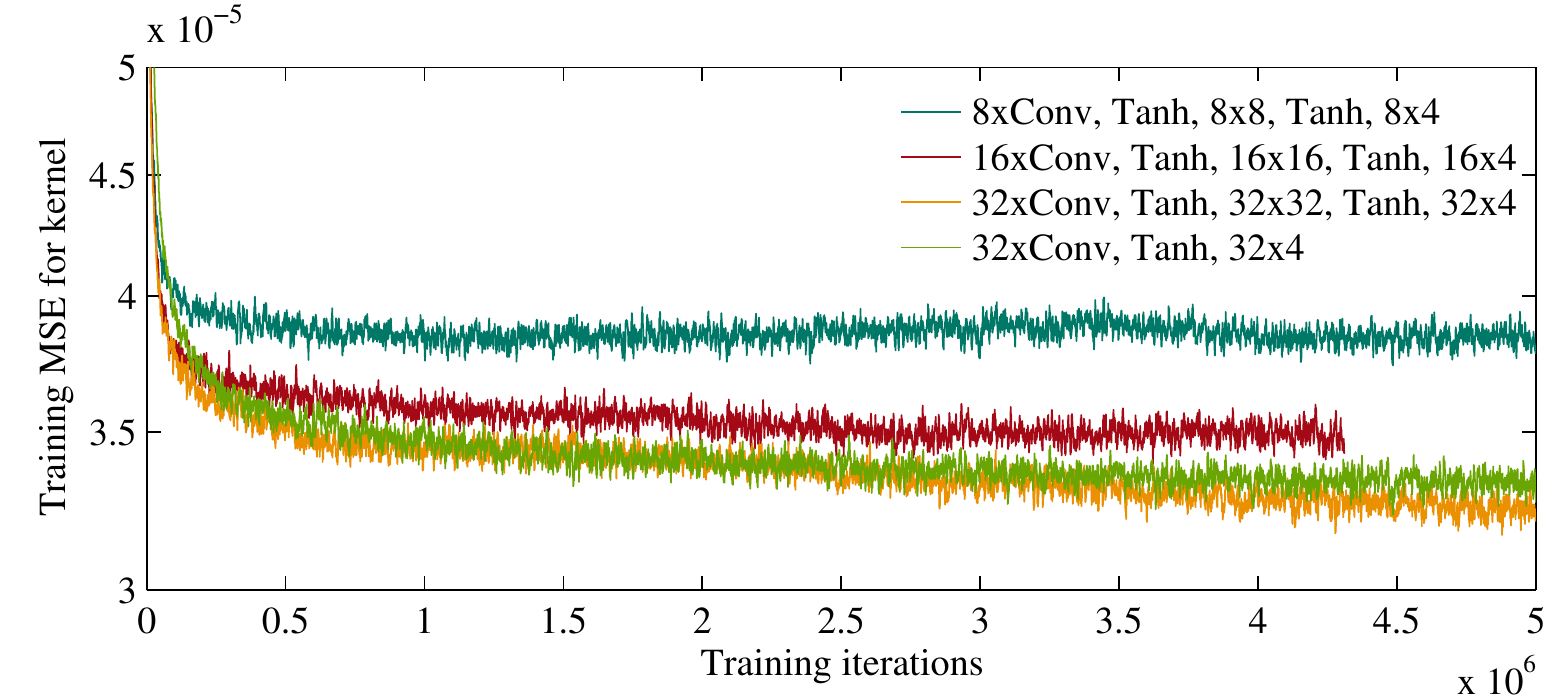}
  \caption[Training curves depending on architecture of a single stage.]{The performance of the network for kernel estimation depends on
the architecture. More filters in the convolutional layer and more hidden
layers are better.}
\label{fig:bn.architecture}
\end{figure}

\begin{figure}[p]
  \centering
  \small
  \includegraphics{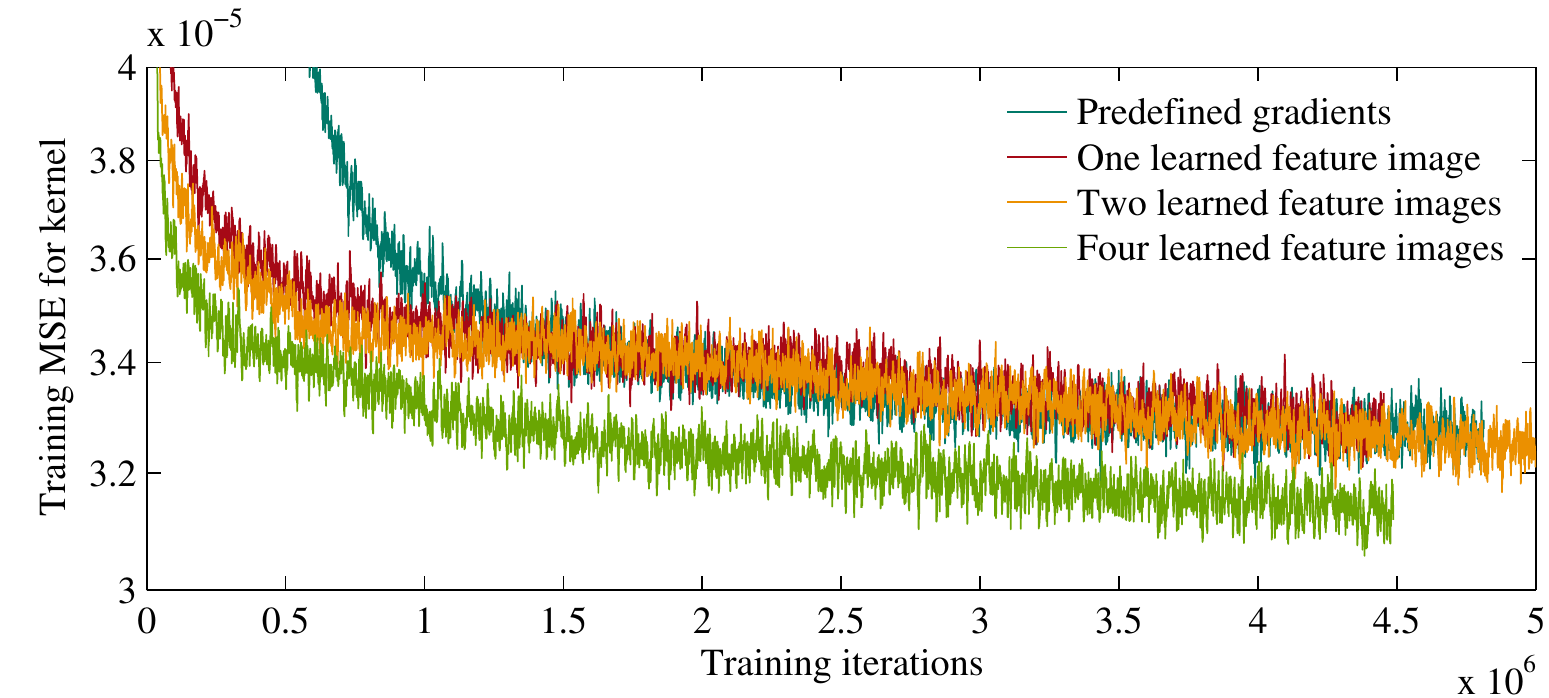}
  \caption[Training curves depending on number of learned gradient-like images.]{The performance of the network depends on the
number of the predicted gradient-like images used in the kernel estimation
module. Predefining $\tilde{\vv{y}}_i$ to x- and y-gradients and
not learning these representations slows down training.}
\label{fig:bn.gradients}
\end{figure}

\begin{figure}[p]
  \centering
  \small
  \includegraphics{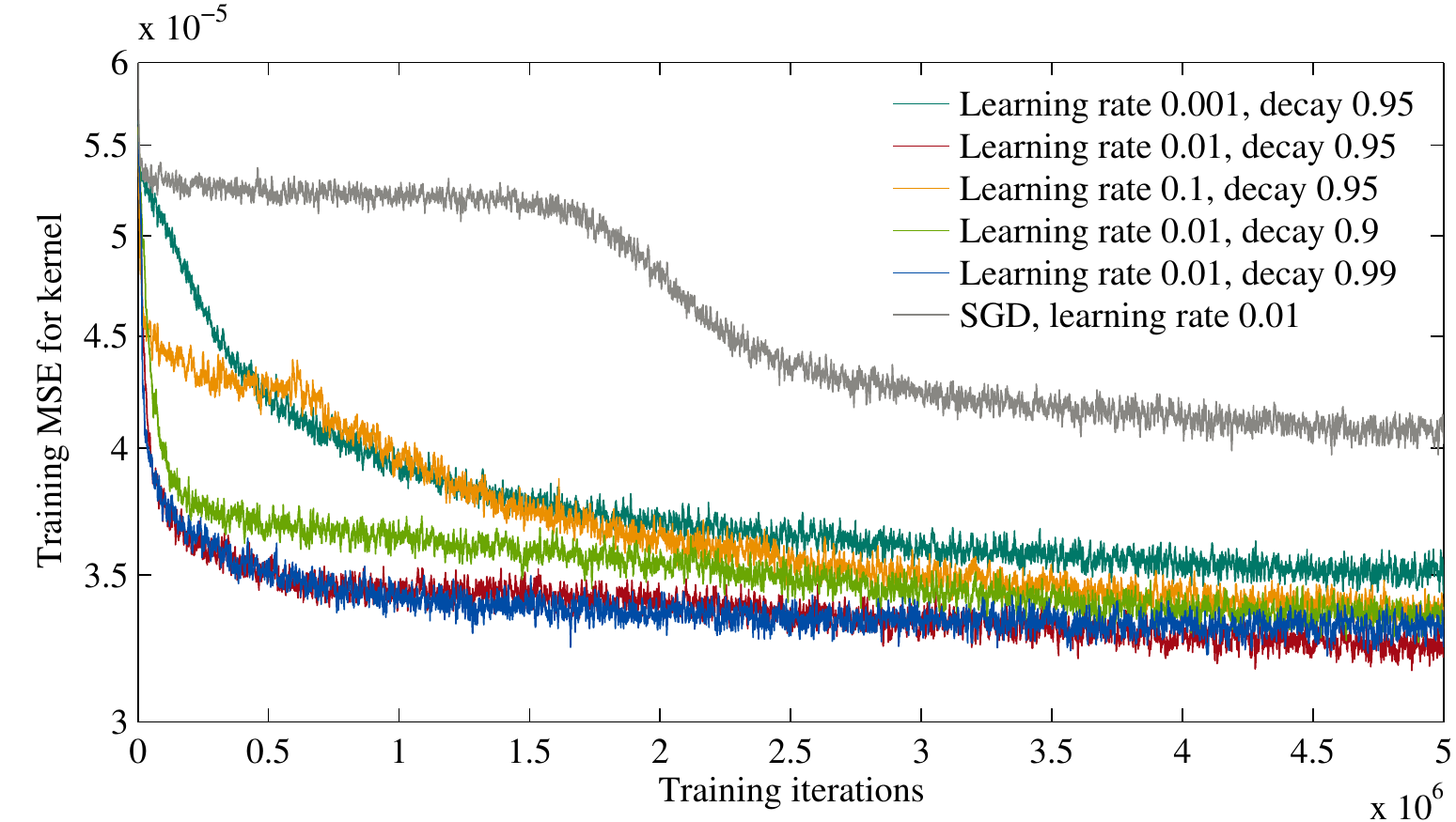}
  \caption[Influence of the learning parameters.]{Influence of the parameters of ADADELTA \autocite{zeiler2012adadelta} or stochastic gradient descent (SGD) on the convergence of the network.}
\label{fig:bn.learning}
\end{figure}

\subsection{Training}

\begin{figure}
  \centering
  \small
  \tightcolsep%
  \begin{tabular}{ccccccccc}
  \includegraphics[width=0.11\textwidth]{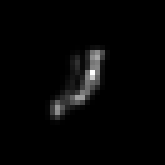}&
  \includegraphics[width=0.11\textwidth]{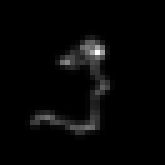}&
  \includegraphics[width=0.11\textwidth]{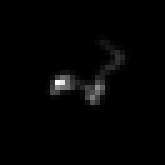}&
  \includegraphics[width=0.11\textwidth]{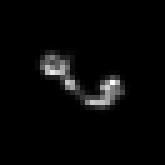}&
  \hfill&
  \includegraphics[width=0.11\textwidth]{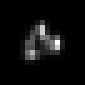}&
  \includegraphics[width=0.11\textwidth]{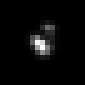}&
  \includegraphics[width=0.11\textwidth]{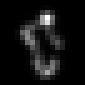}&
  \includegraphics[width=0.11\textwidth]{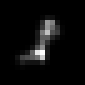}
  \end{tabular}
  \caption[Examples of blurs sampled from a Gaussian process.]{Examples of blurs sampled from a Gaussian process (left: 33\,px, right: 17\,px).}
\label{fig:blurs}
\end{figure}

To train the network, we generate pairs of sharp and blurry images. The sharp
image is sampled from a subset of about 1.6 million images of the ImageNet \autocite{deng2009imagenet}
dataset, randomly cropped to a size of 256$\times$256. Next, a blur trajectory
is generated by sampling both its x- and y-coordinates separately from a
Gaussian Process
\begin{equation}
  f_x(t), f_y(t) \sim \mathcal{GP}\big(0,k(t,t')\big),\;
  k(t,t')=\sigma_f^2\left(1+\frac{\sqrt{5}|t-t'|}{l}+\frac{5{(t-t')}^2}{3l^2}\right)\exp\left(-\frac{\sqrt{5}|t-t'|}{l}\right),
\end{equation}
which is a Matérn covariance function with $\nu=3/2$ \autocite{rasmussen2005gps}.
The length scale $l$ is set to $0.3$, the signal standard deviation $\sigma_f$ to $1/4$.
The
coordinates are then scaled to a fixed blur kernel size and the final kernel is
shifted to have its center of mass in the middle. This simple procedure
generates realistic looking blur kernels, examples for both small and large
kernel sizes are illustrated in Fig.\nobreakspace \ref {fig:blurs}. For every setting, we
generate 1 million noise-free training examples, and add
Gaussian noise during training (by default, $\sigma=0.01$).

To avoid that the training process is disturbed by examples where the blur
kernel cannot be estimated, \eg, a homogeneous area of sky without any gradient
information, we reject examples where less than 6\% pixels have gradients in
x- and y-direction with an absolute value 0.05 or above.

As described in the previous subsection, a network for a certain blur kernel,
\ie, a particular scale, consists of several stages, each iterating between
the \emph{feature extraction}, \emph{kernel estimation} and the \emph{image estimation module}.

We use pre-training for our network: we start by training only one
stage minimizing the L2 error between the estimated and the ground truth
kernel, then add a second stage after about 1 million training steps. For
the next 1000 steps, the parameters of the first stage stay fixed and only
the parameters of the second stage are updated. After that, the network is
trained end-to-end, until a potential next stage is added.

For the update of the parameters, convergence proved to be best with
ADADELTA \autocite{zeiler2012adadelta}, a heuristic weighting scheme of parameter
updates using gradients and updates from previous training steps.
For the influence of its parameters on the convergence speed see
Fig.\nobreakspace \ref {fig:bn.learning}. For our experiments, we choose a learning rate of 0.01
and a decay rate of 0.95.
Moreover, it makes training more robust to outliers with strong
gradients since it divides by the weighted root-mean-square of the seen
gradients, including the current one. Responsible for the mentioned
strong gradients are typically images dominated by abrupt step-edges,
which create ringing artifacts in the
deconvolution Eq.\nobreakspace \textup {(\ref {eq:bn.x_objective})}. In ImageNet these often are
photos of objects with trimmed background. To make the training even more robust,
we don't back-propagate examples with an error above 10 times the
current average error.

\section{Implementation}

We make the code for both training and testing our method available for download. For
training, we use our own C++/CUDA-based neural network toolbox.
After training once
for a certain blur class (\eg, camera shake), which takes about two days per stage, applying the network is very
fast and can be done in Matlab without additional dependencies.

\begin{table}
\small
\centering
\begin{tabular}{c|cccc}
Blur size & 255$\times$255 & 800$\times$800 & 1024$\times$1024 & 2048$\times$2048 \\
\hline
\hline
17$\times$17 & 1.1 & 5.2 & 9.5 & 53.1 \\
25$\times$25 & 1.2 & 14.3 & 18.5 & 89.1 \\
33$\times$33 & 1.6 & 10.7 & 22.6 & 91.9 \\
\end{tabular}
\caption[Run-time for kernel estimation.]{The method is very fast: runtime in seconds for kernel estimation with varying image size on an Intel i5 in Matlab.}
\label{tab:runtime}
\end{table}

The runtimes on an Intel i5 using only Matlab are shown in
Table\nobreakspace \ref {tab:runtime}. The most expensive calculation is the creation of the
multiple hidden representations in the convolutional layer of the NN
(in this example: 32).

\section{Experiments}
\label{sec:experiments}

If not otherwise stated, all experiments were performed with a multi-scale,
triple stage architecture. We use up to three scales for kernels of size
17$\times$17, 25$\times$25, 33$\times$33. On each scale each \emph{feature
extraction module} consists of a convolution layer with 32 filters, a tanh-layer,
a linear recombination to 32 new hidden images, a further tanh-layer, and
a recombination to four gradient-like images, two for $\tilde{\vv{x}}_i$ and
$\tilde{\vv{y}}_i$ each (in the third stage: eight gradient-like images). In the case of
the network
with blur kernels of size 33$\times$33, we deconvolve the estimated kernels with
a small Gaussian with $\sigma=0.5$ to counter the over-smoothing effect of the L2
norm used during training.  For the specific choice of the architecture,
we refer to the influence of model parameters on the kernel estimation
performance in Figs.\nobreakspace  \ref {fig:bn.depth} to\nobreakspace  \ref {fig:bn.architecture} .

\subsection{Image content specific training}
\label{sec:expert.training}

\begin{figure}
\includegraphics[height=2.45cm]{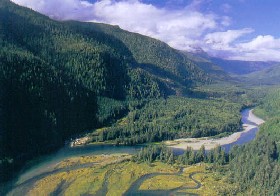}\hspace*{\stretch{1}}
\includegraphics[height=2.45cm]{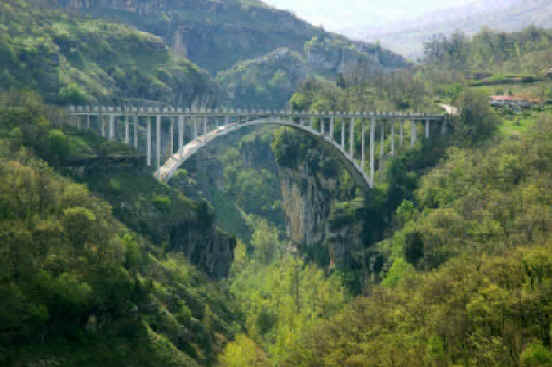}\hspace*{\stretch{1}}
\includegraphics[height=2.45cm]{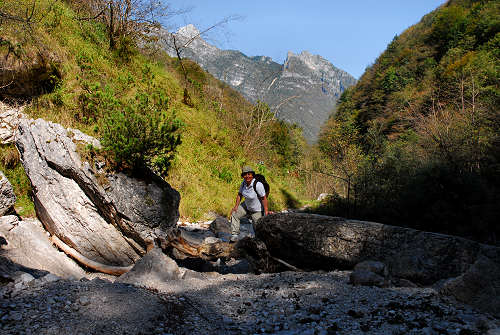}\hspace*{\stretch{1}}
\includegraphics[height=2.45cm]{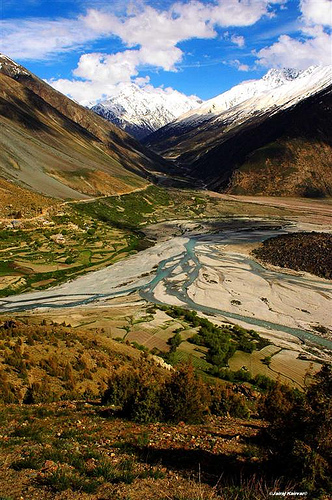}\hspace*{\stretch{1}}
\includegraphics[height=2.45cm]{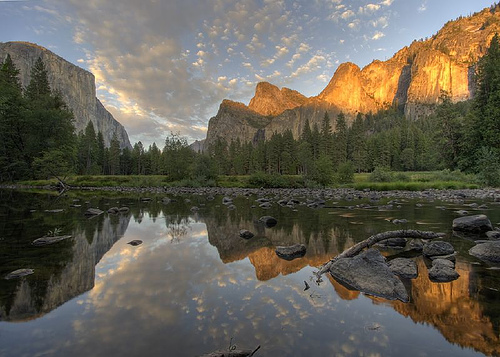} \\
\includegraphics[height=2.73cm]{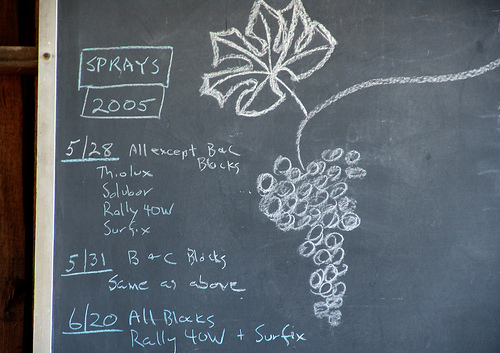}\hspace*{\stretch{1}}
\includegraphics[height=2.73cm]{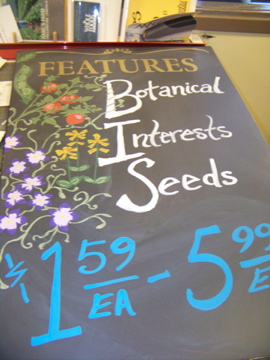}\hspace*{\stretch{1}}
\includegraphics[height=2.73cm]{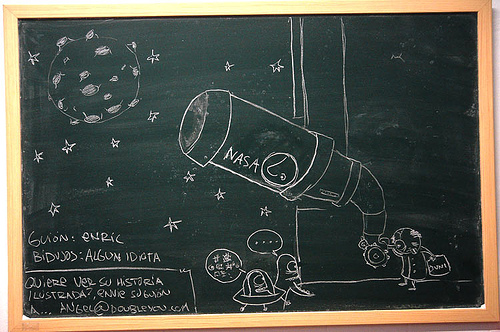}\hspace*{\stretch{1}}
\includegraphics[height=2.73cm]{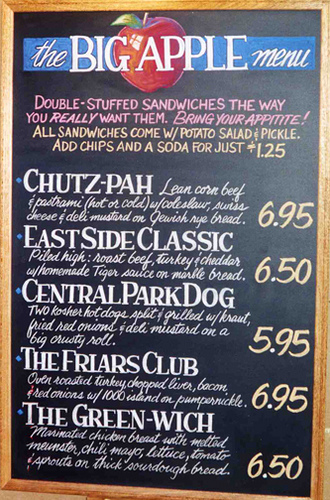}\hspace*{\stretch{1}}
\includegraphics[height=2.73cm]{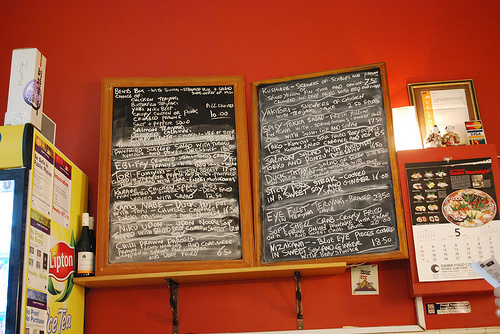}
\caption[Typical example images of \emph{valley} and
  \emph{blackboard} categories from ImageNet used for
  content specific training.]{Typical example images of \emph{valley} (top row) and
  \emph{blackboard} (bottom row) categories from ImageNet used for
  content specific training.}
\label{fig:class.examples}
\end{figure}

\begin{figure}
\tightcolsep%
\small
\centering
\begin{tabular}{cccc}
\begin{tikzpicture}[every node/.style={anchor=south east,inner sep=0pt}]
\node (fig1) at (0,0){\includegraphics[width=0.98\quadpw]{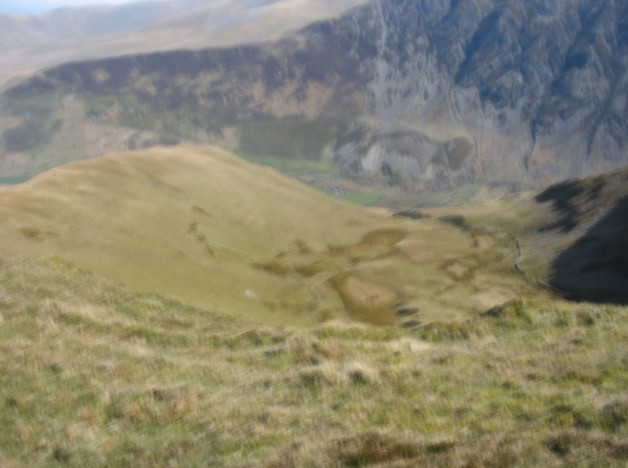}};%
\node (fig2) at (0.02\quadpw,-0.1\quadpw){\includegraphics[width=0.3\quadpw]{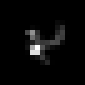}};%
\end{tikzpicture}
&
\begin{tikzpicture}[every node/.style={anchor=south east,inner sep=0pt}]
\node (fig1) at (0,0){\includegraphics[width=0.98\quadpw]{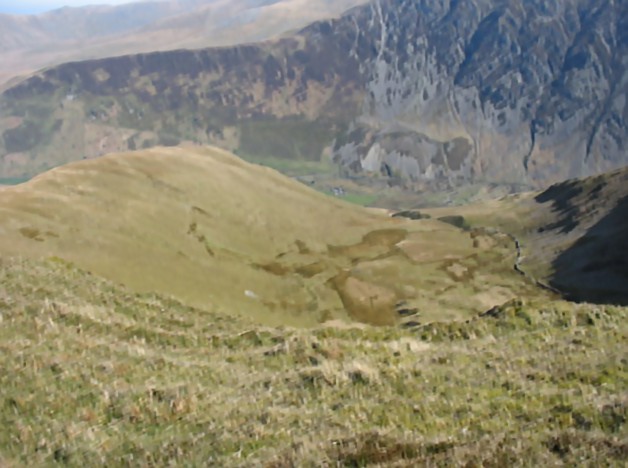}};%
\node (fig2) at (0.02\quadpw,-0.1\quadpw){\includegraphics[width=0.3\quadpw]{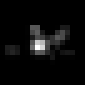}};%
\end{tikzpicture}
&
\begin{tikzpicture}[every node/.style={anchor=south east,inner sep=0pt}]
\node (fig1) at (0,0){\includegraphics[width=0.98\quadpw]{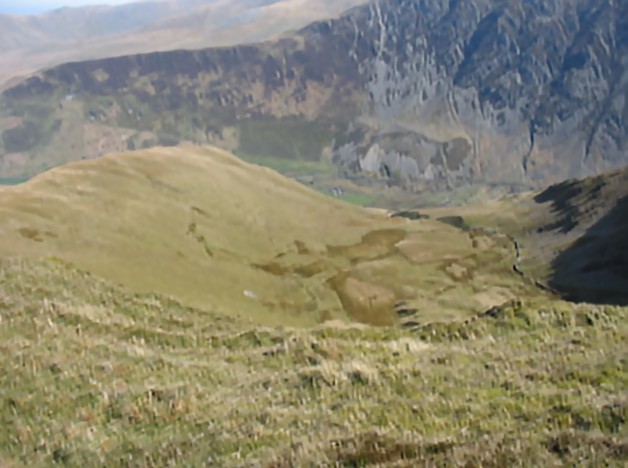}};%
\node (fig2) at (0.02\quadpw,-0.1\quadpw){\includegraphics[width=0.3\quadpw]{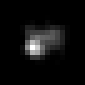}};%
\end{tikzpicture}
&
\begin{tikzpicture}[every node/.style={anchor=south east,inner sep=0pt}]
\node (fig1) at (0,0){\includegraphics[width=0.98\quadpw]{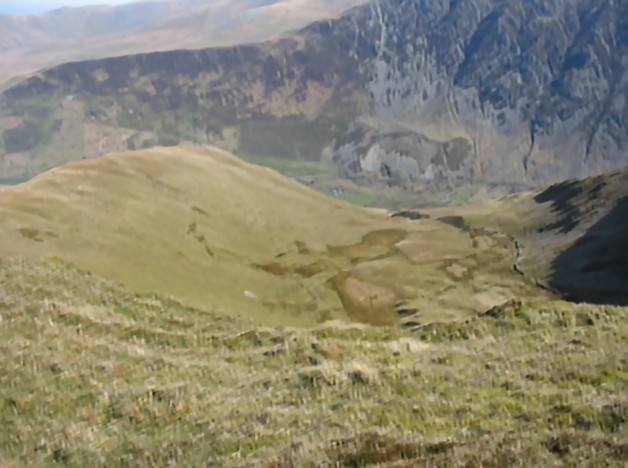}};%
\node (fig2) at (0.02\quadpw,-0.1\quadpw){\includegraphics[width=0.3\quadpw]{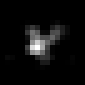}};%
\end{tikzpicture}
\\[3ex]
\begin{tikzpicture}[every node/.style={anchor=south east,inner sep=0pt}]
\node (fig1) at (0,0){\includegraphics[width=0.98\quadpw]{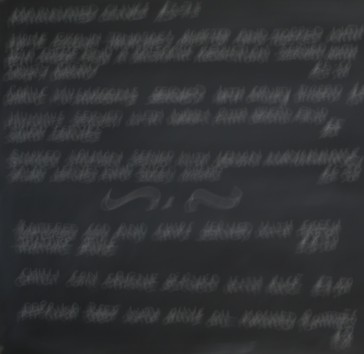}};%
\node (fig2) at (0.02\quadpw,-0.1\quadpw){\includegraphics[width=0.3\quadpw]{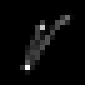}};%
\end{tikzpicture}
&
\begin{tikzpicture}[every node/.style={anchor=south east,inner sep=0pt}]
\node (fig1) at (0,0){\includegraphics[width=0.98\quadpw]{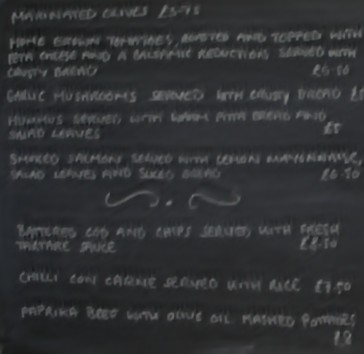}};%
\node (fig2) at (0.02\quadpw,-0.1\quadpw){\includegraphics[width=0.3\quadpw]{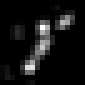}};%
\end{tikzpicture}
&
\begin{tikzpicture}[every node/.style={anchor=south east,inner sep=0pt}]
\node (fig1) at (0,0){\includegraphics[width=0.98\quadpw]{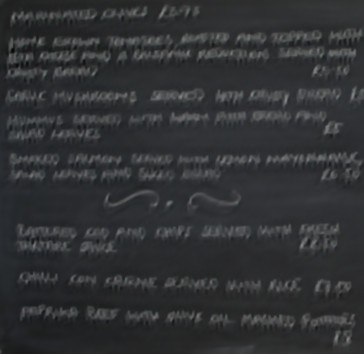}};%
\node (fig2) at (0.02\quadpw,-0.1\quadpw){\includegraphics[width=0.3\quadpw]{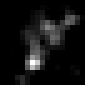}};%
\end{tikzpicture}
&
\begin{tikzpicture}[every node/.style={anchor=south east,inner sep=0pt}]
\node (fig1) at (0,0){\includegraphics[width=0.98\quadpw]{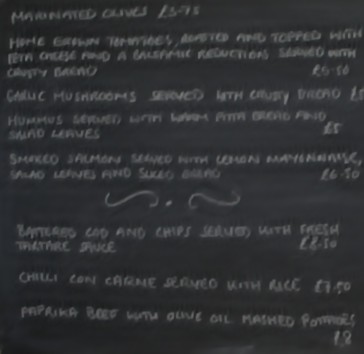}};%
\node (fig2) at (0.02\quadpw,-0.1\quadpw){\includegraphics[width=0.3\quadpw]{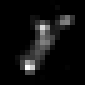}};%
\end{tikzpicture}
\\
Blurry image with & Deblurring result of  & Deblurring result with & Deblurring result with \\
ground truth kernel & \textcite{xu_cvpr2013} & content \emph{agnostic} training & content \emph{specific} training
\end{tabular}
\caption[Comparison of deblurring results for NNs that
  have been trained with image examples from the entire ImageNet
  dataset (content \emph{agnostic}) and from particular subsets
  (content \emph{specific}).]{Comparison of deblurring results for NNs that
  have been trained with image examples from the entire ImageNet
  dataset (content \emph{agnostic}) and from particular subsets
  (content \emph{specific}), \ie, image categories \emph{valley} (top
  row) and \emph{blackboard} (bottom row). We also show the results of
  the state-of-the-art method \autocite{xu_cvpr2013}.}
\label{fig:expertnet}
\end{figure}

A number of recent works \autocite{hu_eccv2012,sun2013,wang2013} have
pointed out the shortcoming of state-of-the-art algorithms
\autocite{cho2009fast,Xu2010} to depend on the presence of strong salient
edges and their diminished performance in the case of images that
contain textured scenes such as natural landscape images. The reason
for this is the deficiency of the so-called image prediction step,
which applies a combination of bilateral and shock filtering to restore
latent edges that are used for subsequent kernel estimation.

In this context, learning the latent image prediction step offers a
great advantage: by training our network with a particular class of
images, it is able to focus on those features that are informative for
the particular type of image. In other words, the network learns
\emph{content-specific} nonlinear filters, which yield improved performance.

To demonstrate this, we used the same training procedure as described
above, however, we reduced the training set to images from a specific
image category within the ImageNet dataset. In particular, we used the
image category \emph{valley}\footnote{ImageNet 2011 Fall Release $>$
  Geological formation, formation $>$ Natural depression, depression
  $>$ Valley, vale
  (\url{http://www.image-net.org/synset?wnid=n09468604})} containing a
total of 1395 pictures. In a second experiment, we trained a network
on the image category \emph{blackboard}\footnote{ImageNet 2011 Fall
  Release $>$ Artifact, artifact $>$ Sheet, flat solid $>$ Blackboard,
  chalkboard (\url{http://www.image-net.org/synset?wnid=n02846511})}
with a total of 1376 pictures. Figure\nobreakspace \ref {fig:class.examples}
shows typical example images from these two classes.
Fig.\nobreakspace \ref {fig:expertnet} compares deblurring results of the
state-of-the-art algorithm described in \autocite{xu_cvpr2013} with our
approach trained on images sampled from the entire ImageNet
dataset, and trained on the aforementioned image categories only. Note that the
particular images shown in Fig.\nobreakspace \ref {fig:expertnet} were not used
for training. We see that content specific training outperforms both content
agnostic training and the generic state-of-the-art method.

\begin{figure}[t]
\tightcolsep%
\small
\begin{tabular}{cccc}
\begin{tikzpicture}[every node/.style={anchor=south east,inner sep=0pt}]
\node (fig1) at (0,0){\includegraphics[width=0.98\quadpw]{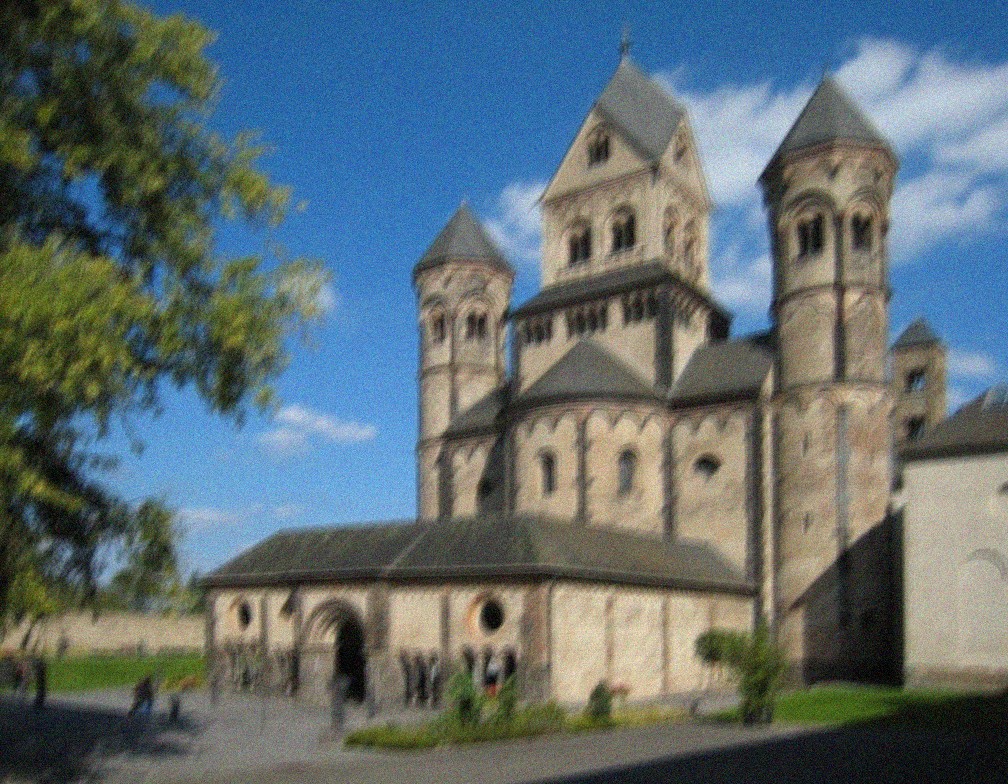}};%
\node (fig2) at (0.02\quadpw,-0.1\quadpw){\includegraphics[width=0.3\quadpw]{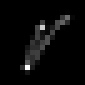}};%
\end{tikzpicture}
&
\begin{tikzpicture}[every node/.style={anchor=south east,inner sep=0pt}]
\node (fig1) at (0,0){\includegraphics[width=0.98\quadpw]{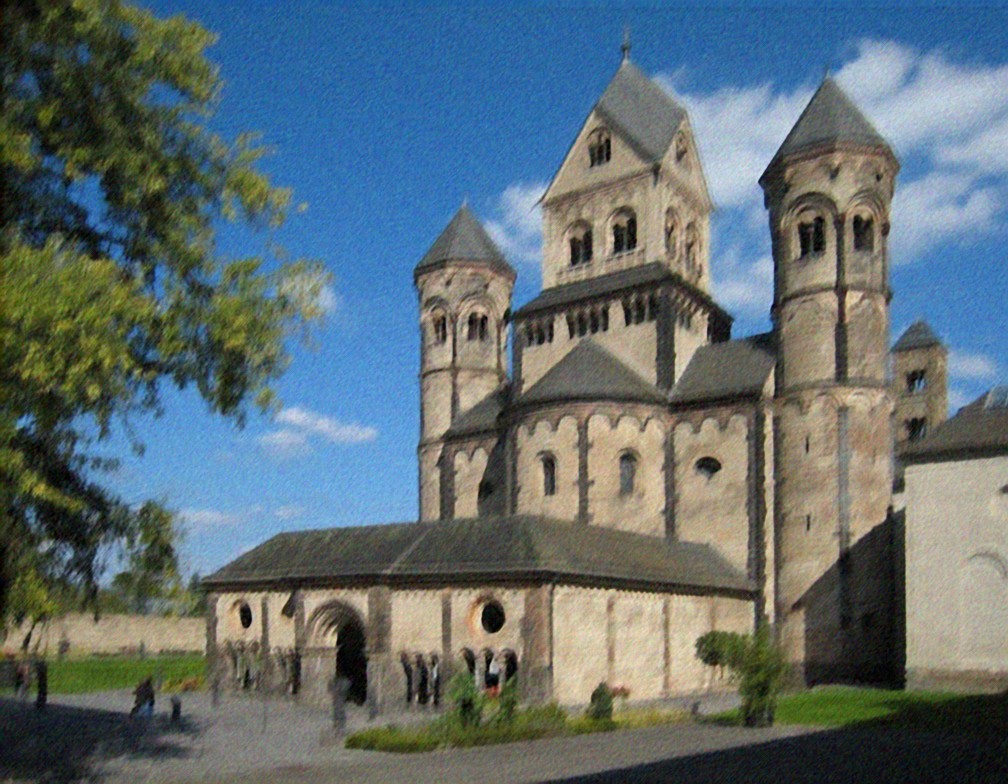}};%
\node (fig2) at (0.02\quadpw,-0.1\quadpw){\includegraphics[width=0.3\quadpw]{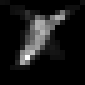}};%
\end{tikzpicture}
&
\begin{tikzpicture}[every node/.style={anchor=south east,inner sep=0pt}]
\node (fig1) at (0,0){\includegraphics[width=0.98\quadpw]{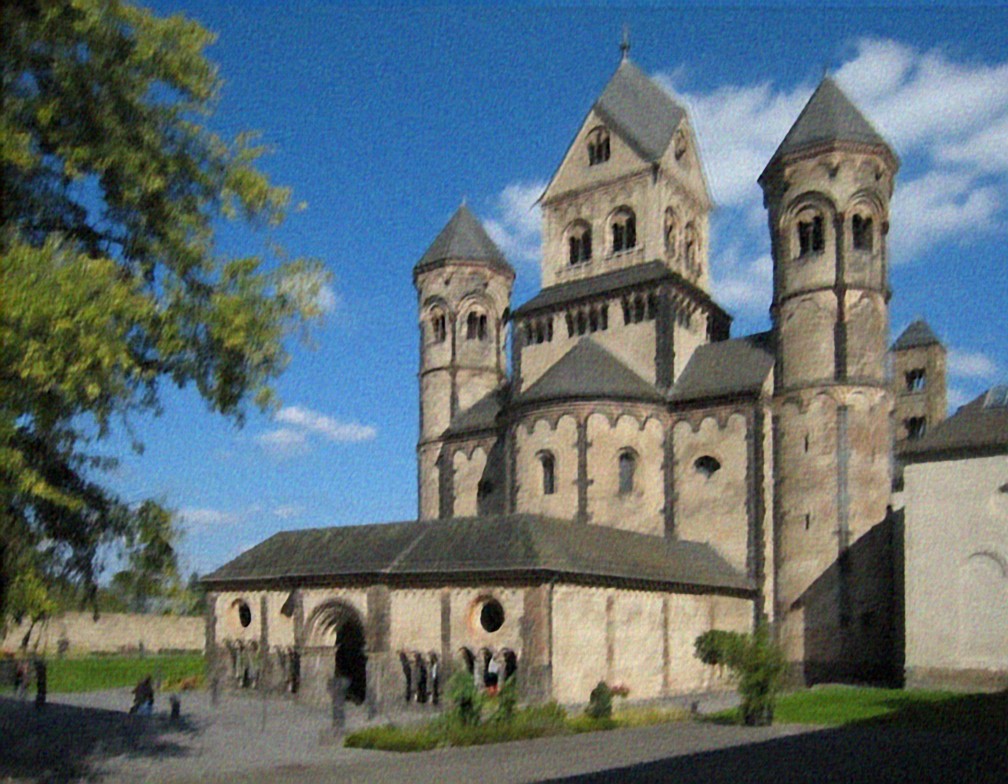}};%
\node (fig2) at (0.02\quadpw,-0.1\quadpw){\includegraphics[width=0.3\quadpw]{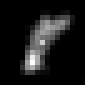}};%
\end{tikzpicture}
&
\begin{tikzpicture}[every node/.style={anchor=south east,inner sep=0pt}]
\node (fig1) at (0,0){\includegraphics[width=0.98\quadpw]{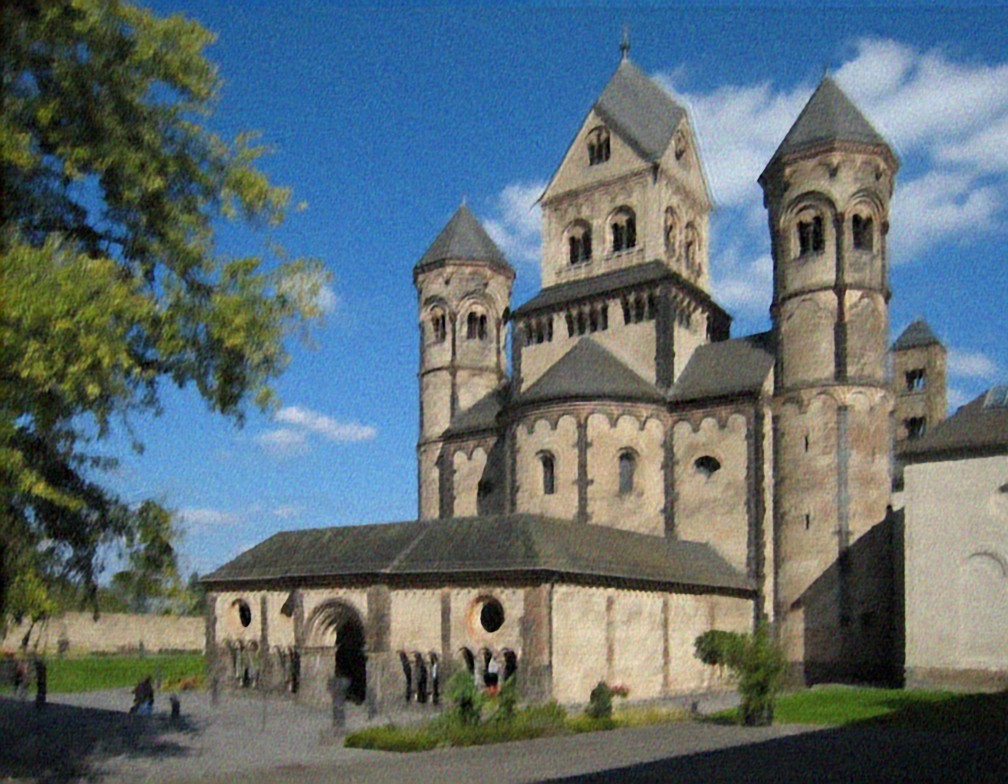}};%
\node (fig2) at (0.02\quadpw,-0.1\quadpw){\includegraphics[width=0.3\quadpw]{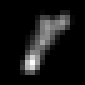}};%
\end{tikzpicture}
\\
Blurry image with & Result
of \autocite{zhong_cvpr2013} & Deblurring result w.& Deblurring result w.\\
ground truth kernel & PSNR 23.17 & noise \emph{agnostic} training & noise
\emph{specific} training \\
& & PSNR 23.29 & \textbf{PSNR 23.41} 
\end{tabular}
\caption[Comparison of deblurring results for NNs that
  have been trained with different amounts of noise added to the sample
  images during training.]{Comparison of deblurring results for NNs that
  have been trained with different amounts of noise added to the sample
  images during training. The network that has been trained with the
  same amount of noise as the input blurry image (5\% noise)
  performs best. We also show the results of a recently proposed
  deblurring method tailored for increased levels of noise
  \autocite{zhong_cvpr2013}.}
\label{fig:noise}
\end{figure}

\subsection{Noise specific training}
\label{sec:noise.training}

Typically, image noise impedes kernel estimation. To counter noise in
blurry images, current state-of-the-art deblurring algorithms apply a
denoising step during latent image prediction such as bilateral
filtering \autocite{cho2009fast} or Gaussian filtering
\autocite{Xu2010}. However, in a recent work \autocite{zhong_cvpr2013}, the
authors show that for increased levels of noise current methods fail
to yield satisfactory results and propose a novel robust deblurring
algorithm. Again, if we include image noise in our training phase, our
network is able to adapt and learn filters that perform better in the
presence of noise. In particular, we trained a network on images with
Gaussian noise of 5\% added during the training phase. Figure\nobreakspace \ref {fig:expertnet} compares the results for an image taken from
\autocite{zhong_cvpr2013} with 5\% Gaussian noise for a network trained with 1\%
and 5\% of added Gaussian noise during training, respectively. We also
show the result of \autocite{zhong_cvpr2013} and compare peak signal-to-noise ratio (PSNR) for objective
evaluation. All results use the same non-blind deconvolution of
\autocite{zhong_cvpr2013}. The noise specific training is most successful, but
even the noise agnostic NN outperforms the non-learned method
on this example.

\subsection{Spatially-varying blur}
\label{sec:space-variant}

\begin{figure}
\tightcolsep%
\small
\begin{tabular}{ccc}
\includegraphics[width=\triplepw]{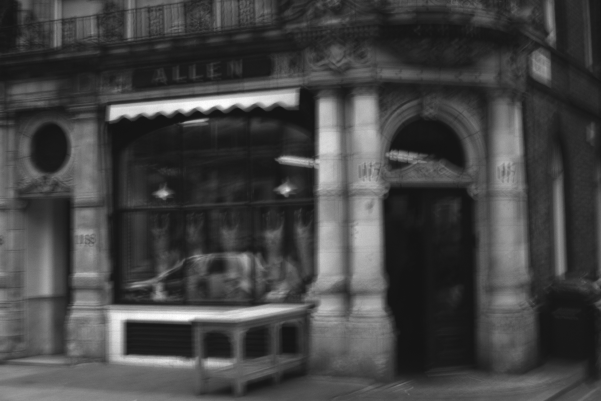} &
\includegraphics[width=\triplepw]{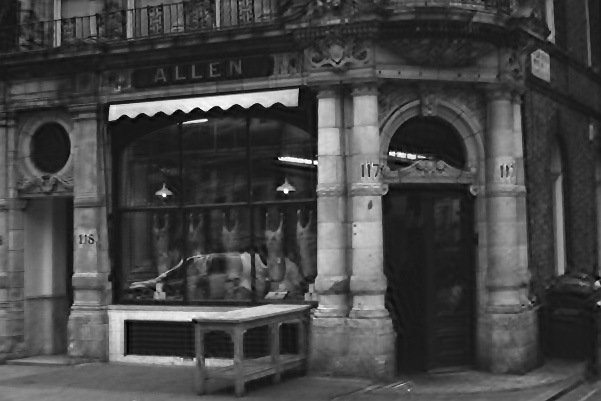} &
\includegraphics[width=\triplepw]{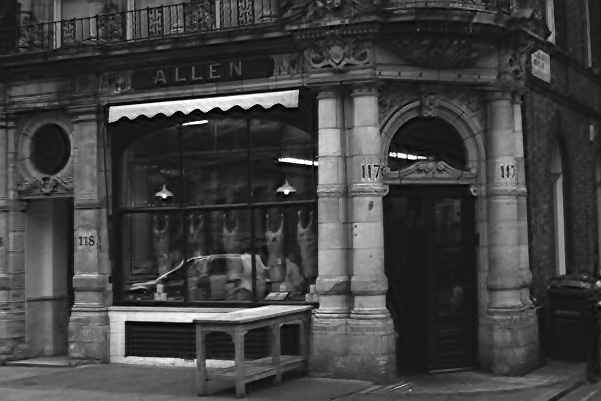}\\
Blurry image & \textcite{harmeling_nips2010} & \textcite{hirsch_iccv2011}\\[1.5ex]
\includegraphics[width=\triplepw]{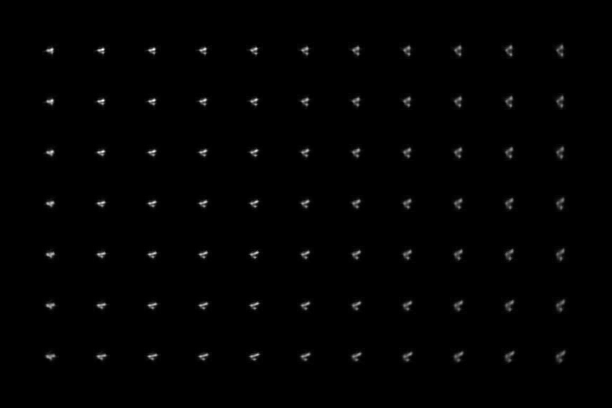}&
\includegraphics[width=\triplepw]{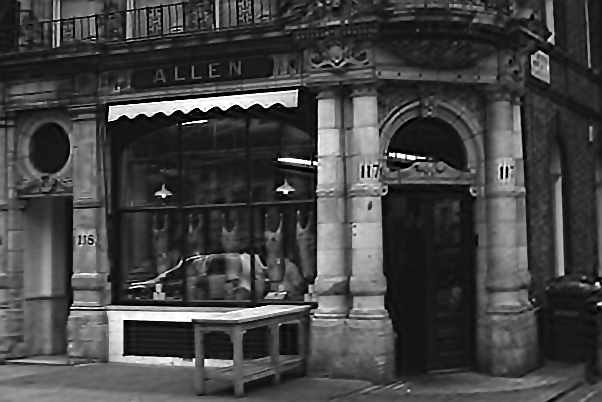}&
\includegraphics[width=\triplepw]{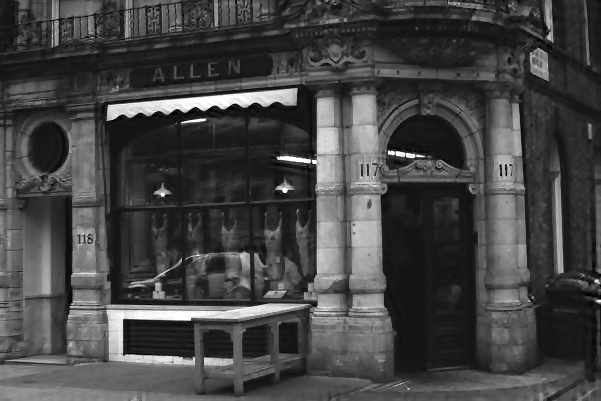} \\
Estimated PSF & Ours & \textcite{xu_cvpr2013} \\
\end{tabular}
\caption[Comparison on \emph{Butcher Shop} example 
  of state-of-the-art deblurring methods for
  removing non-uniform blur together with our estimated PSF.]{Comparison on \emph{Butcher Shop} example 
  \autocite{harmeling_nips2010} of state-of-the-art deblurring methods for
  removing non-uniform blur together with our estimated PSF.}
\label{fig:spatially}
\end{figure}

\begin{figure}
\centering
\small
\tightcolsep%
\begin{tabular}{ccc}
\includegraphics[width=\triplepw]{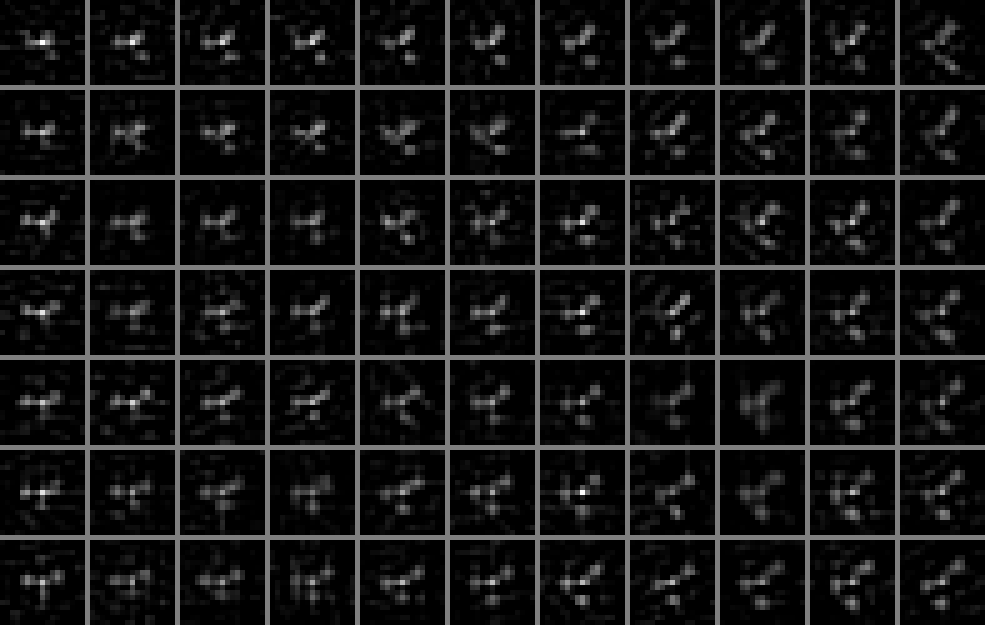}&
\includegraphics[width=\triplepw]{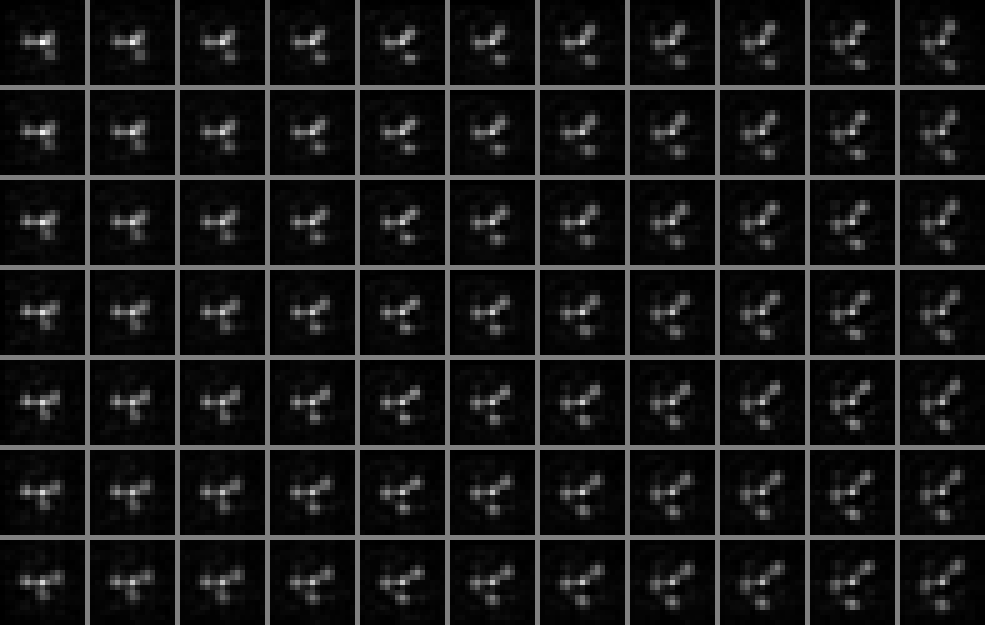}&
\includegraphics[width=\triplepw]{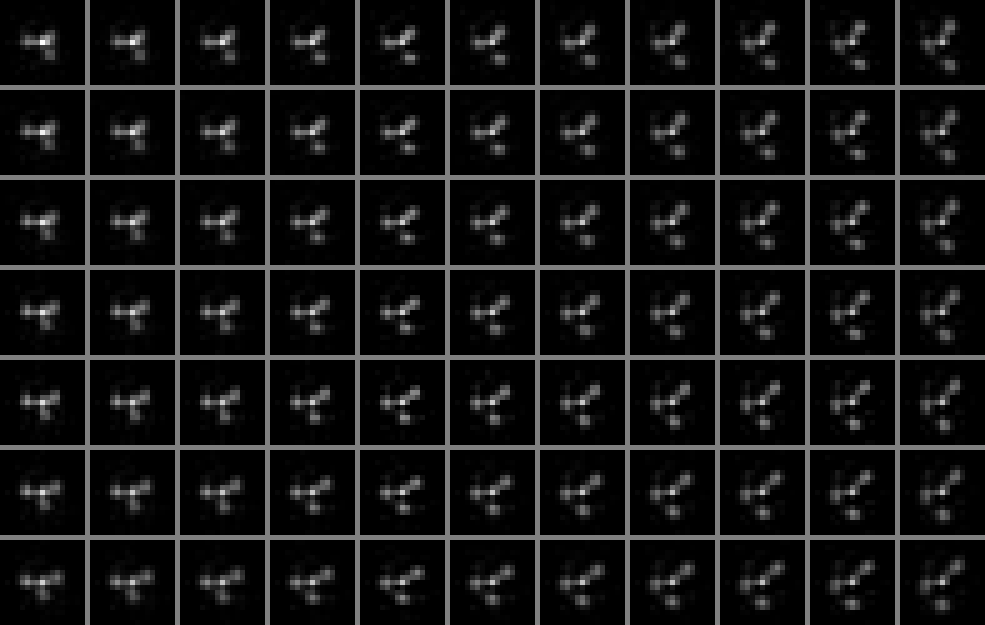}\\
Result of direct kernel estimation& Result after basis projection & Result after basis projection \\
 &  & and subsequent sparsification
\end{tabular}
\caption[Visualisation of kernel estimation in the case of
  spatially-varying blur for the \emph{Butcher Shop} example.]{Visualisation of our kernel estimation in the case of
  spatially-varying blur for the \emph{Butcher Shop} example shown in
  Fig.\nobreakspace \ref {fig:spatially}. The left panel shows the
  kernel estimated with Eq.\nobreakspace \textup {(\ref {eq:bn.direct})}, the middle and right panels show the
  kernel after applying a subsequent projection step to our motion
  basis, \ie the result of Eq.\nobreakspace \textup {(\ref {eq:bn.project})} with $\eta$ set to 1.0
  and 0.0314 respectively.}
\label{fig:bn.space_variant_kernel}
\end{figure}

Since the prediction step of our trained NN is independent
of the convolution model, we can also use it in conjunction with the
recently proposed fast forward model of \autocite{hirsch_iccv2011} to
restore images with spatially-varying blur. To this end, we replace
the objective function Eq.\nobreakspace \textup {(\ref {eq:bn.x_objective})} with Eq.~(8) of
\autocite{hirsch_iccv2011} in our kernel estimation module in a network
trained for spatially invariant deconvolution. We solve for $k$
in a two-step procedure: first we compute local blur kernels using the
efficient filter flow (EFF) model of \autocite{hirsch_cvpr2010}; in a second step
we project the blur kernels onto a motion basis aka
\autocite{hirsch_iccv2011}, as explained below. Figure\nobreakspace \ref {fig:spatially} shows a
comparison between recent state-of-the-art algorithms for
spatially-varying blur along with our deblurring result that features
comparable quality.

For the estimation of spatially-varying blur we solve the following
objective
\begin{align}
  \sum_i \|\tilde{X}_i \, \tilde{\vv{k}}  - \tilde{\vv{y}}_i\|^2 + \beta_k \|\tilde{\vv{k}}\|^2
  \label{eq:bn.k_objective_spatially}
\end{align}
in our kernel estimation module. Here $\tilde{X}_i$ denotes the
EFF matrix of $\tilde{\vv{x}}$ (\cf Eq.~(9) in
\autocite{hirsch_cvpr2010}) and $\tilde{\vv{k}}$ the stacked sequence of local
kernels $\tilde{\vv{k}}^{(r)}$, one for each patch that are enumerated by
index $r$. Since Eq.\nobreakspace \textup {(\ref {eq:bn.k_objective_spatially})} is quadratic in
$\tilde{\vv{k}}$, we can solve for a local blur $\tilde{\vv{k}}^{(r)}$ in a single step
\begin{align}
  \tilde{\vv{k}}^{(r)}_{direct} \approx \F^\H \frac{\sum_i \conj{\F C_r \Diag(\vv{w}^{(r)}) \, \tilde{\vv{x}}_i} \odot (\F C_r \Diag(\vv{w}^{(r)}) \, \tilde{\vv{y}}_i)}%
  {\sum_i \vert \F C_r \Diag(\vv{w}^{(r)}) \, \tilde{\vv{x}}_i\vert^2 + \beta_k},
  \label{eq:bn.direct}
\end{align}
where $C_r$ are appropriately chosen cropping matrices, and $\vv{w}^{(r)}$
are window functions matching the size of $\tilde{\vv{x}}_i$ and $\tilde{\vv{y}}_i$.
Note that Eq.\nobreakspace \textup {(\ref {eq:bn.direct})} is only approximately true and is
motivated by Eq.~(8) in \autocite{hirsch_iccv2011}. Subsequently, we
project the estimated kernel computed by Eq.\nobreakspace \textup {(\ref {eq:bn.direct})} to a
motion blur kernel basis. In our experiments we use the same basis as
\autocite{hirsch_iccv2011} comprising translations within the image plane
and in-plane rotations only. This additional projection step
constrains the estimated blur to physically plausible ones. Formally,
we compute
\begin{align}
  \tilde{\vv{k}}^{(r)}_{est} \approx B^{(r)} \, \, T_\eta \, \underbrace{\sum_r {(B^{(r)})}^T \tilde{\vv{k}}^{(r)}_{direct}}_{\VV{\upmu}},
  \label{eq:bn.project}
\end{align}
where again we make use of the notation chosen in
\autocite{hirsch_iccv2011}, \ie $B^{(r)}$ denotes the motion blur kernel
basis for patch $r$. Then $\VV{\upmu}$ are the coefficients in the basis
of valid motion blurs. $T_\eta$ denotes a thresholding operator that
sets all elements to zero below a certain threshold whereby the
threshold is chosen such that only $\eta$ percent of entries remain
non-zero. This thresholding step is motivated by \autocite{cho2009fast}, who
also apply a hard thresholding step to the estimated kernels in order
to get rid of spurious
artefacts. Figure\nobreakspace \ref {fig:bn.space_variant_kernel} shows the
intermediate results of our kernel estimation procedure in the case of
spatially-varying blur.

\subsection{Comparisons}
\subsubsection{Benchmark Datasets}

\begin{figure}[t]
\tightcolsep%
\small
{\centering
\includegraphics{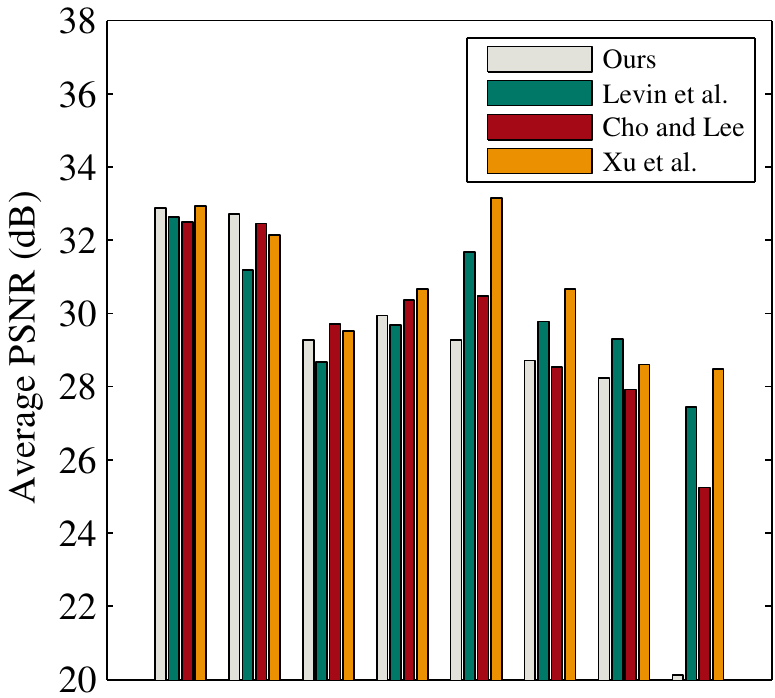}\includegraphics{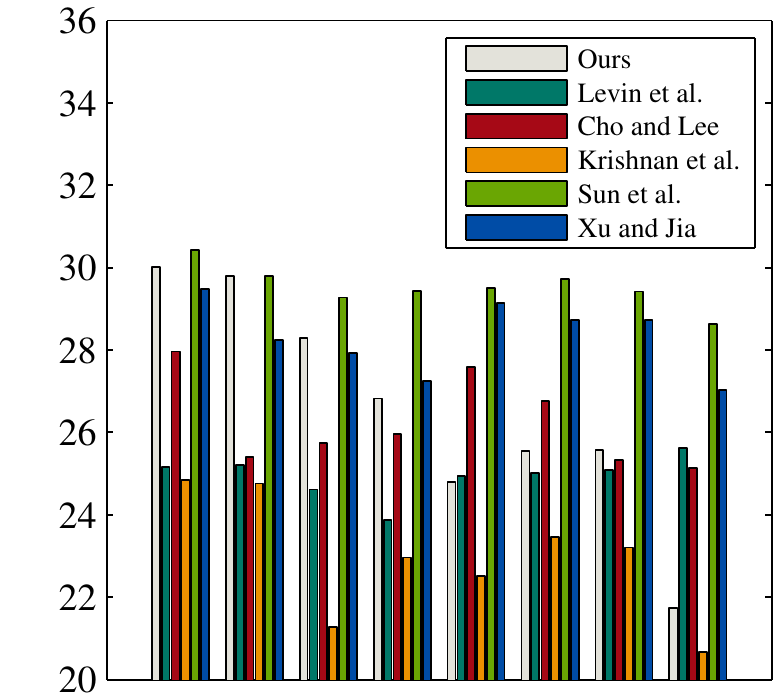}}
\begin{tabular}{ccccccccc}
\hspace*{13.3mm}&
\includegraphics[width=0.04\textwidth]{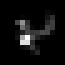} &
\includegraphics[width=0.04\textwidth]{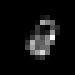} &
\includegraphics[width=0.04\textwidth]{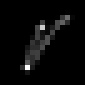} &
\includegraphics[width=0.04\textwidth]{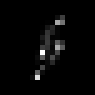} &
\includegraphics[width=0.04\textwidth]{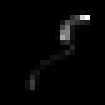} &
\includegraphics[width=0.04\textwidth]{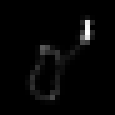} &
\includegraphics[width=0.04\textwidth]{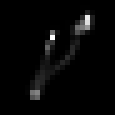} &
\includegraphics[width=0.04\textwidth]{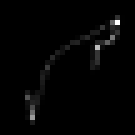}
\end{tabular}\hspace{3.4mm}
\begin{tabular}{ccccccccc}
\hspace*{13.3mm}&
\includegraphics[width=0.04\textwidth]{levin_blurs/k5} &
\includegraphics[width=0.04\textwidth]{levin_blurs/k3} &
\includegraphics[width=0.04\textwidth]{levin_blurs/k2} &
\includegraphics[width=0.04\textwidth]{levin_blurs/k1} &
\includegraphics[width=0.04\textwidth]{levin_blurs/k6} &
\includegraphics[width=0.04\textwidth]{levin_blurs/k7} &
\includegraphics[width=0.04\textwidth]{levin_blurs/k8} &
\includegraphics[width=0.04\textwidth]{levin_blurs/k4}
\end{tabular}
\caption[Results on the benchmark dataset of Levin et al. and the extended benchmark of Sun et al.]{Results of the benchmark dataset of \textcite{levin2009understanding} and the extended benchmark of \textcite{sun2013}. The results are sorted according to blur kernel
  size. While for kernels up to a size of 25$\times$25 pixels, our approach
  yields comparable results, it falls short for larger blur
  kernels. Reasons for the performance drop are discussed in
  Section\nobreakspace \ref {sec:failure}.}
\label{fig:benchmark}
\end{figure}

We evaluate our method on the standard test sets from~\autocite{levin2009understanding,sun2013}.
The four images of~\autocite{levin2009understanding} are 255$\times$255 pixels in size and are
artificially blurred each with eight different blur kernels and contain 1\% additive
Gaussian noise. The performance is illustrated in Fig.\nobreakspace \ref {fig:benchmark} on
the left, with blur kernels sorted according to increasing size. We compare with
\textcite{levin2011efficient}, \textcite{cho2009fast}, and \textcite{xu_cvpr2013}. While our method is competitive with the
state of the art for small blur kernels, our method falls short in performance
for blur kernel sizes above 25$\times$25 pixels. We discuss reasons for this in
Section\nobreakspace \ref {sec:failure}.
The second benchmark from~\autocite{sun2013}
extends this dataset to 80 new images with about one megapixel in size each, using
the same blur kernels as in~\autocite{levin2009understanding}.  Results are shown
in Fig.\nobreakspace \ref {fig:benchmark} on the right. Here we compare with \textcite{levin2011efficient}, \textcite{cho2009fast}, \textcite{krishnan2011normalized}, \textcite{sun2013}, and \textcite{Xu2010}, where however~\autocite{sun2013} has runtime in the
order of hours.
Again, we see competitive performance for small blur kernels.

\subsubsection{Real-World Images}
\label{sec:realworld}

\begin{figure}
\tightcolsep%
\small
\begin{tabular}{ccc}
\includegraphics[width=\triplepw]{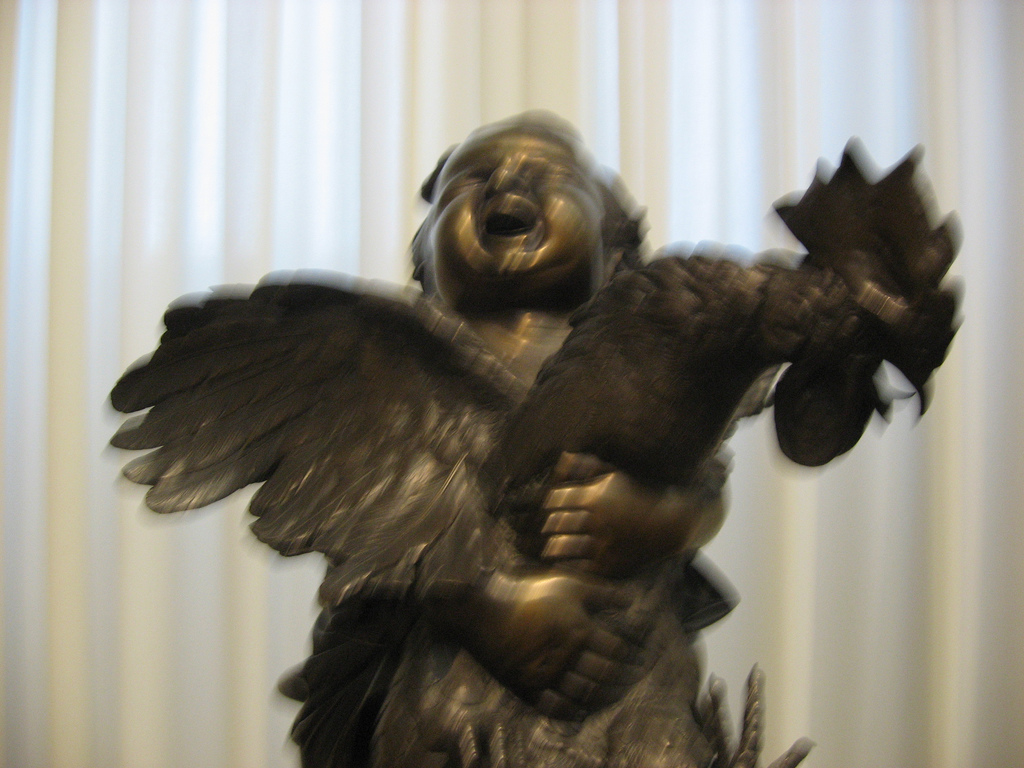} &
\includegraphics[width=\triplepw]{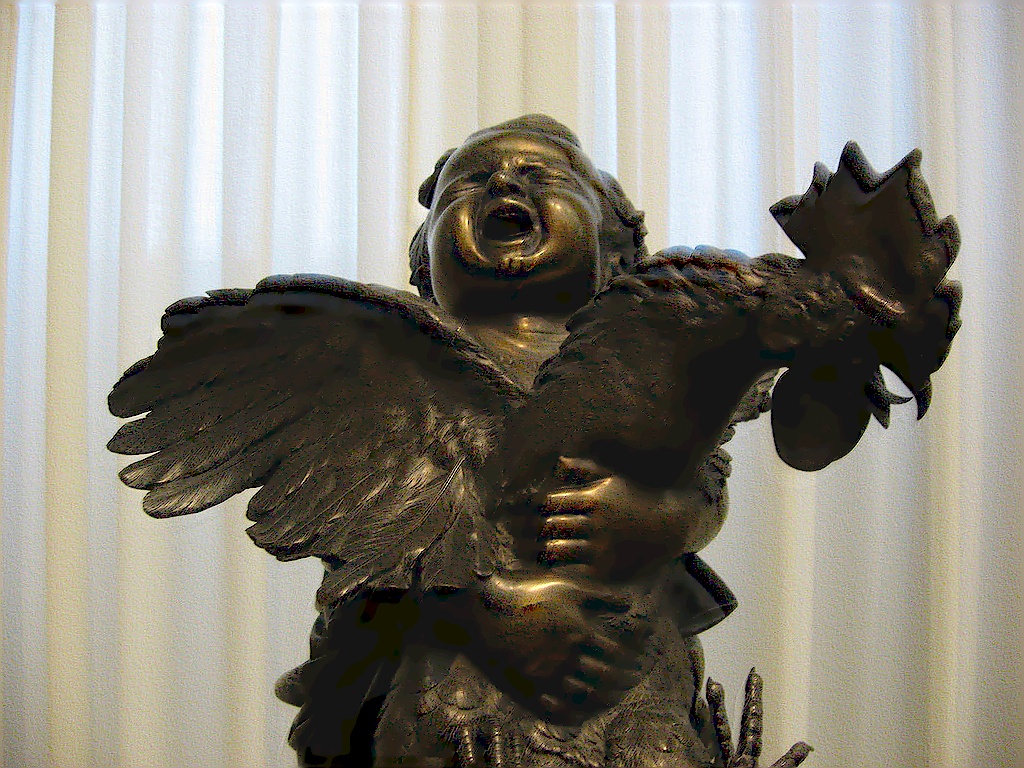} &
\includegraphics[width=\triplepw]{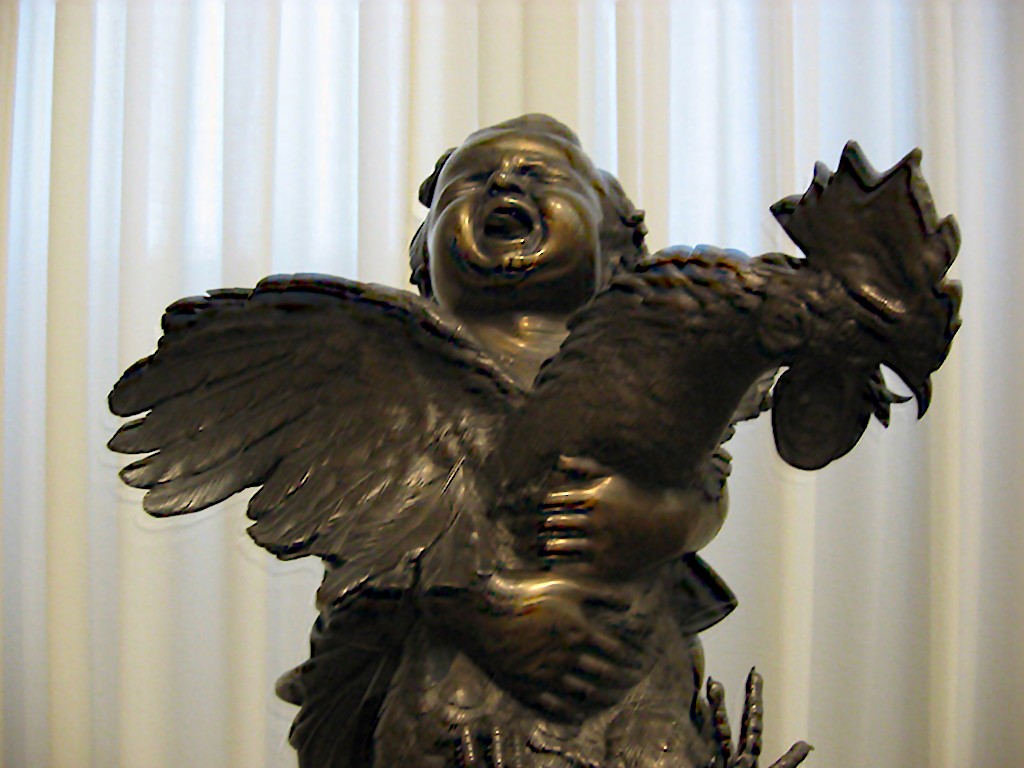} \\
Blurry image & Cho et Lee \autocite{cho2009fast} & Our results \\[1.5ex]
\includegraphics[width=\triplepw]{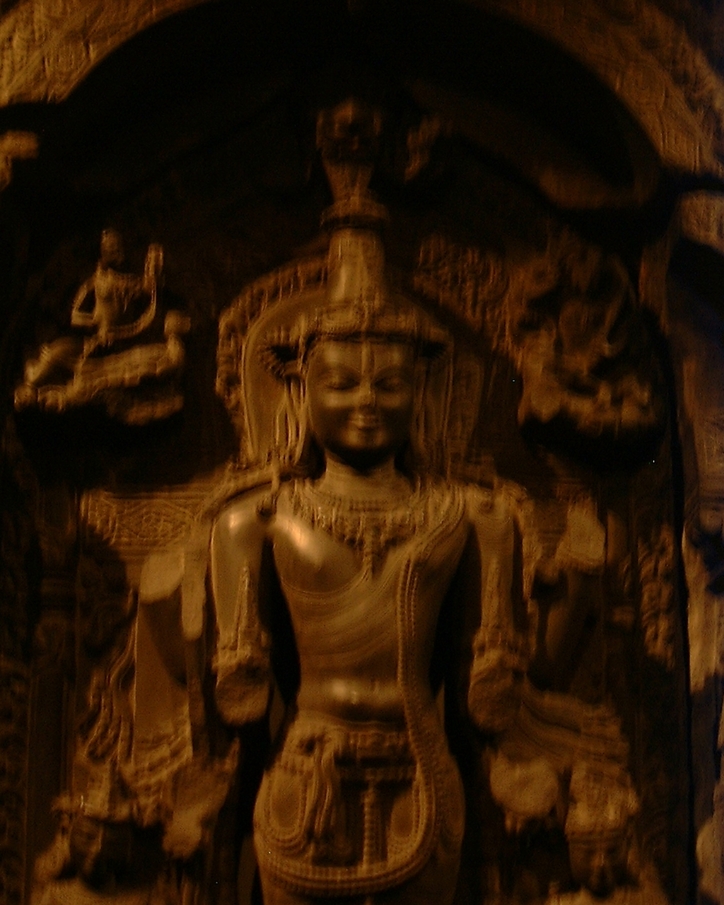}&
\includegraphics[width=\triplepw]{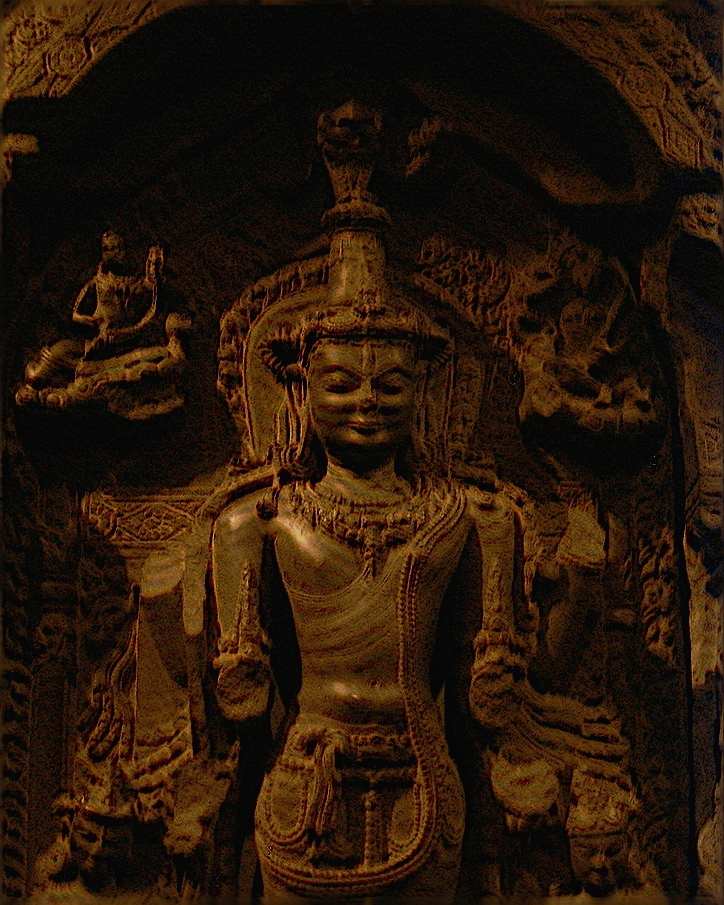}&
\includegraphics[width=\triplepw]{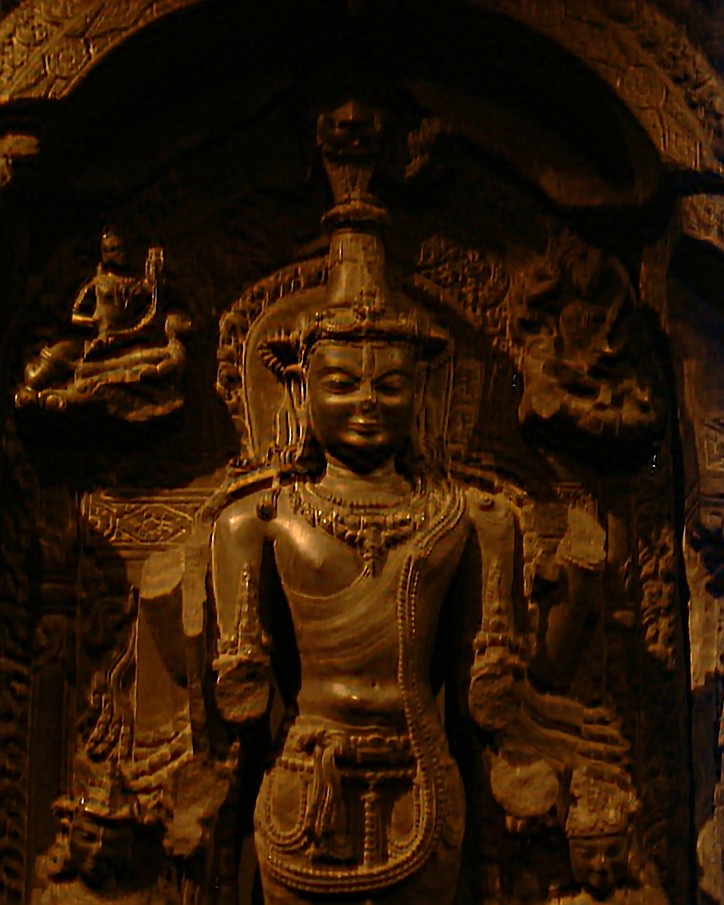}\\
Blurry image & \Textcite{Fergus2006} & Our result \\[1.5ex]
\includegraphics[width=\triplepw]{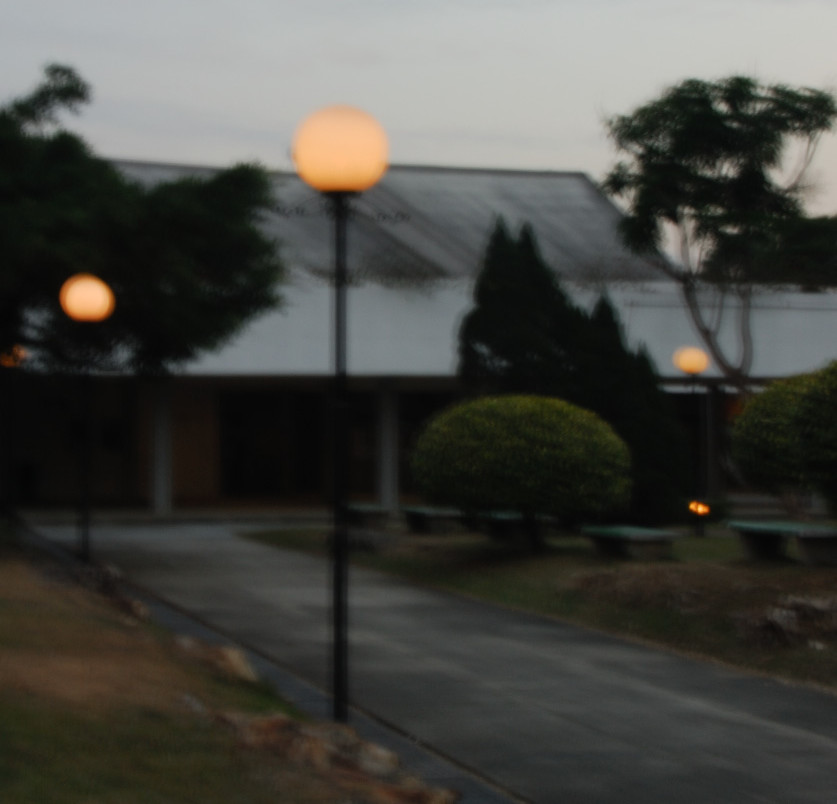}&
\includegraphics[width=\triplepw]{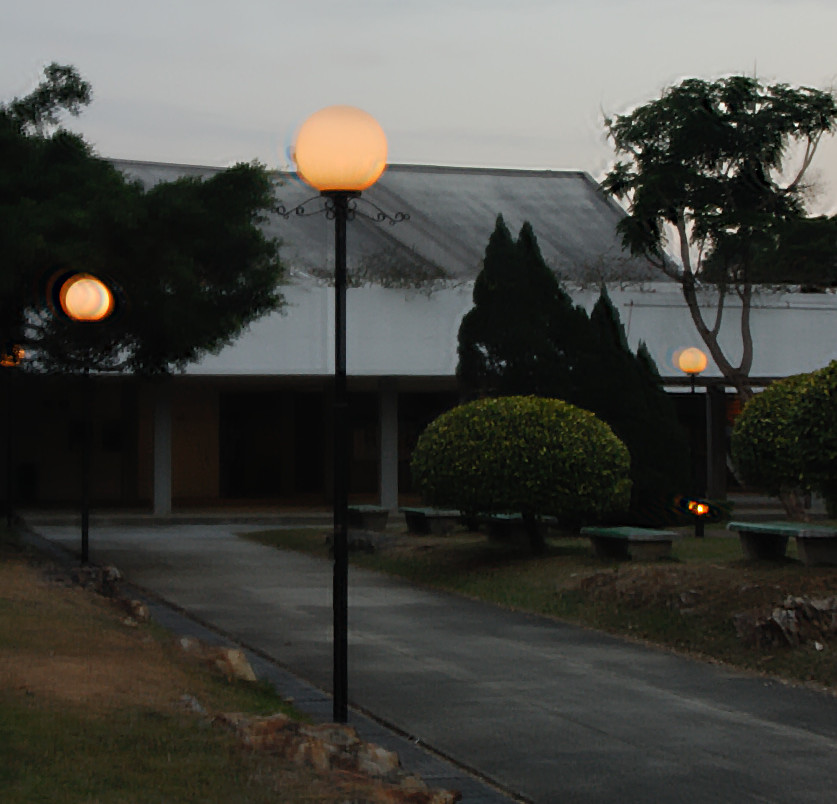}&
\includegraphics[width=\triplepw]{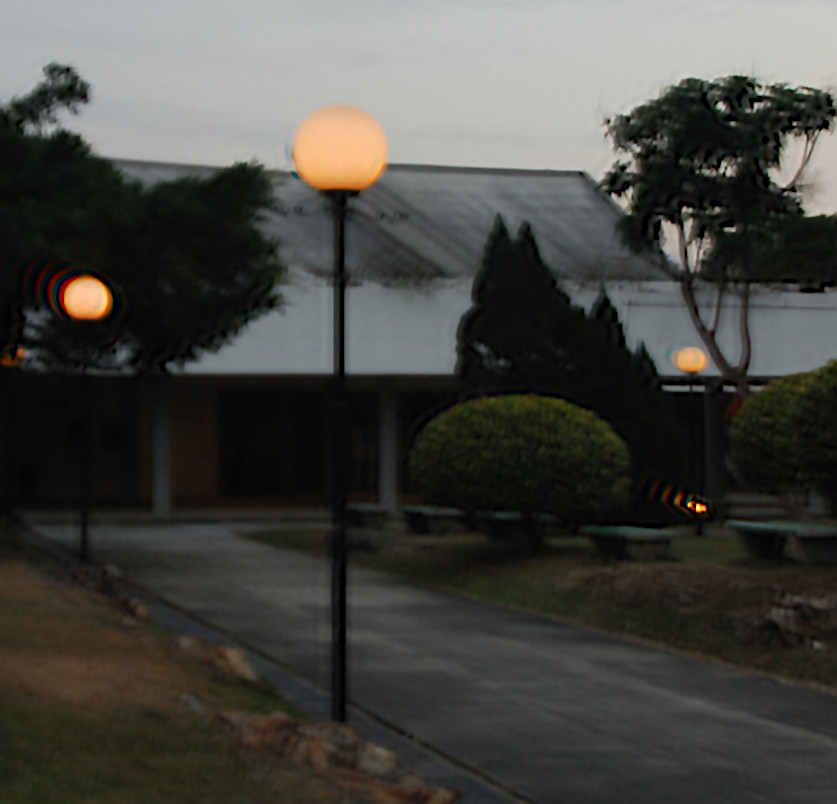}\\
Blurry image & \Textcite{shan2008hqdeblurring} & Our result
\end{tabular}
\caption{Comparison on real-world
  example images taken from the literature with spatially invariant blur.}
\label{fig:bn.realworld_invariant}
\end{figure}

\begin{figure}
\tightcolsep%
\small
\begin{tabular}{ccc}
\includegraphics[width=\triplepw]{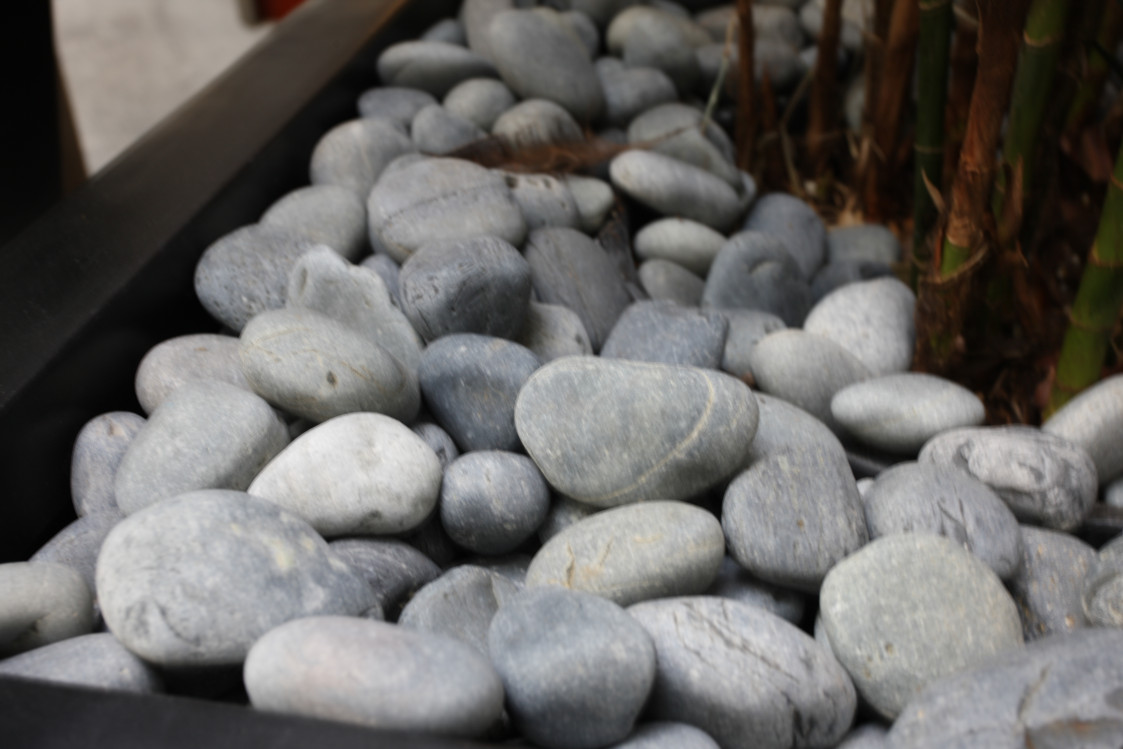}&
\includegraphics[width=\triplepw]{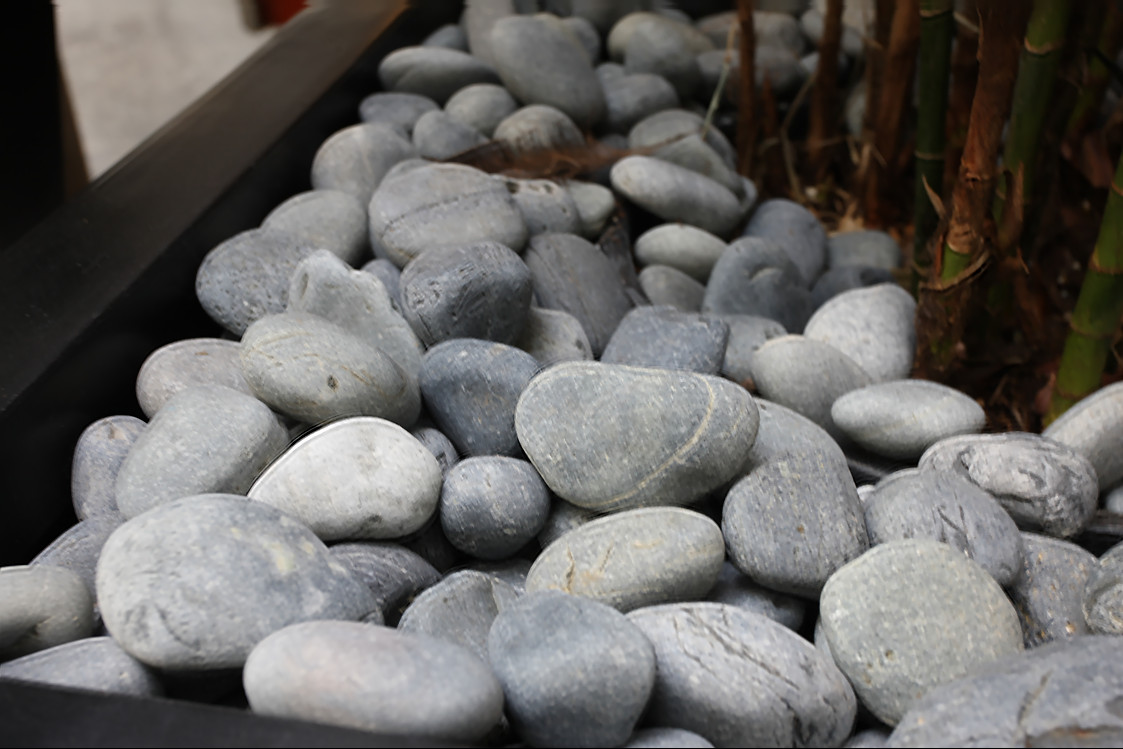}&
\includegraphics[width=\triplepw]{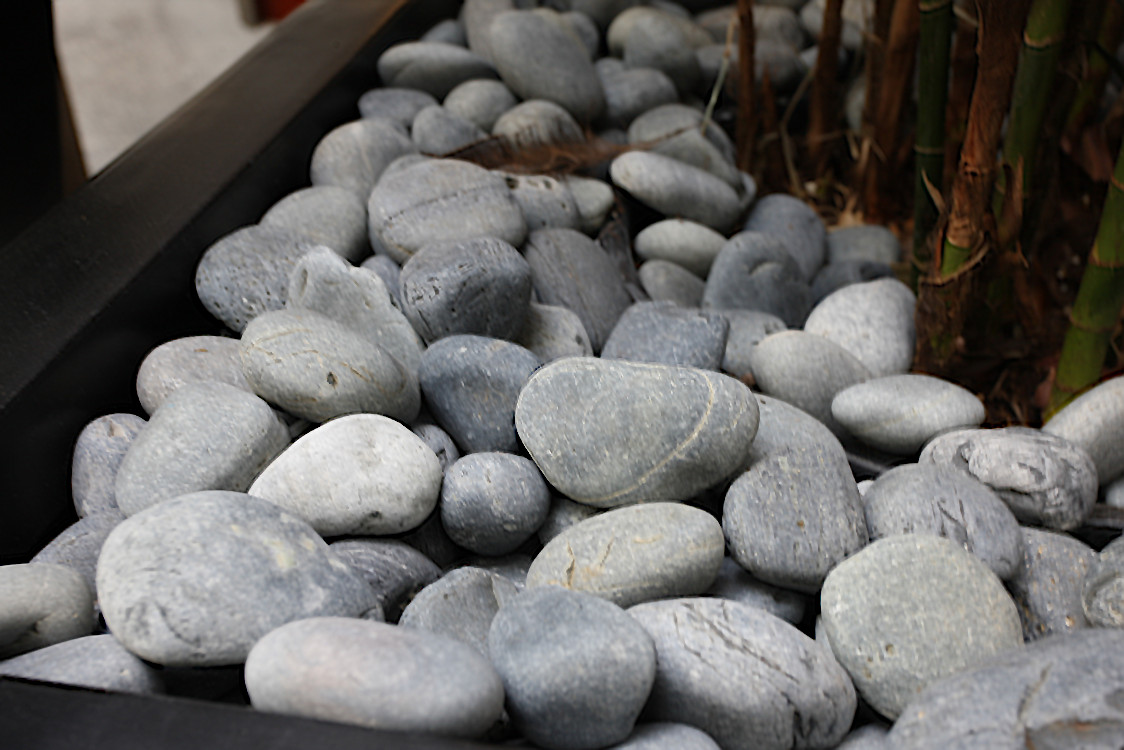} \\
Blurry image & \textcite{joshi_2010} & Our result \\[1.5ex]
\includegraphics[width=\triplepw]{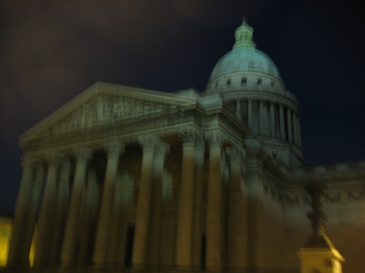}&
\includegraphics[width=\triplepw]{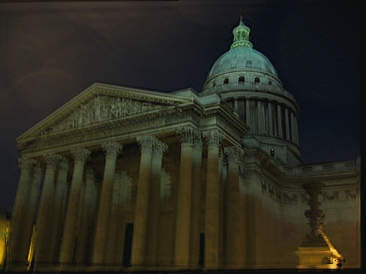}&
\includegraphics[width=\triplepw]{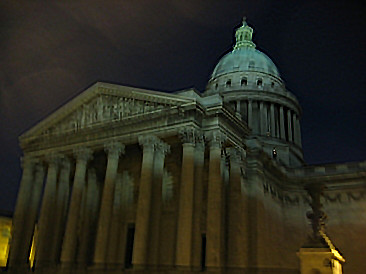}\\
Blurry image \autocite{whyte2010nonunifrom} & \Textcite{whyte2010nonunifrom} & Our result\\[1.5ex]
\includegraphics[width=\triplepw]{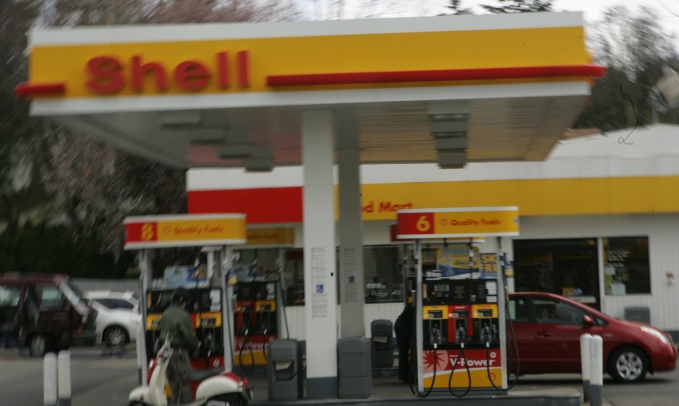}&
\includegraphics[width=\triplepw]{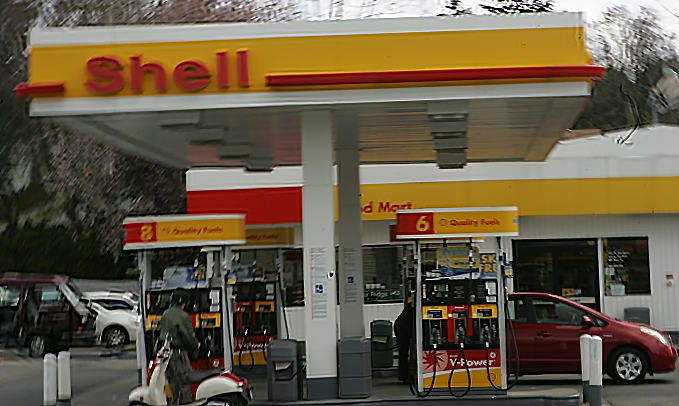}&
\includegraphics[width=\triplepw]{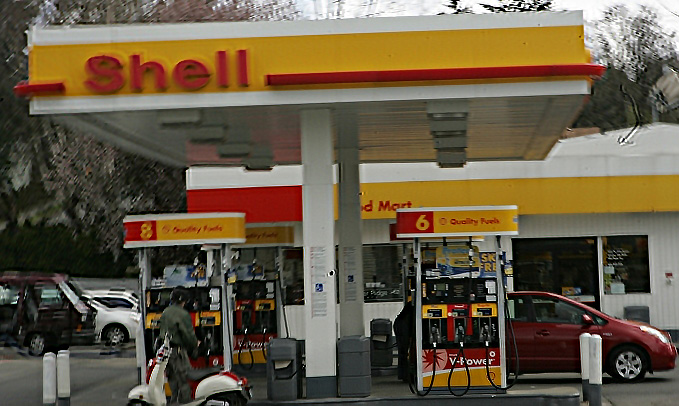}\\
Blurry image \autocite{gupta2010deblurring} & \Textcite{gupta2010deblurring} & \Textcite{hirsch_iccv2011}\\[1.5ex]
&
\includegraphics[width=\triplepw]{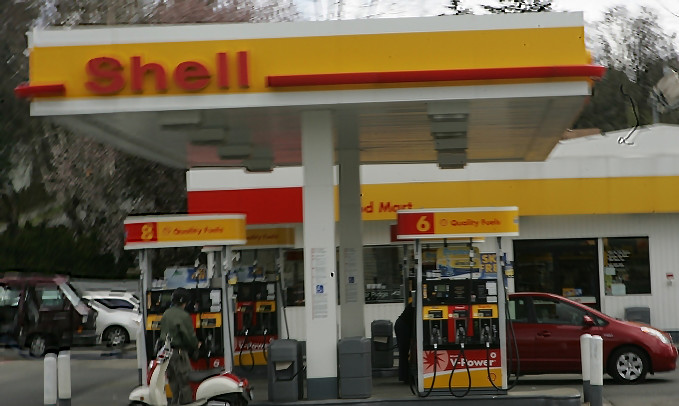}&
\includegraphics[width=\triplepw]{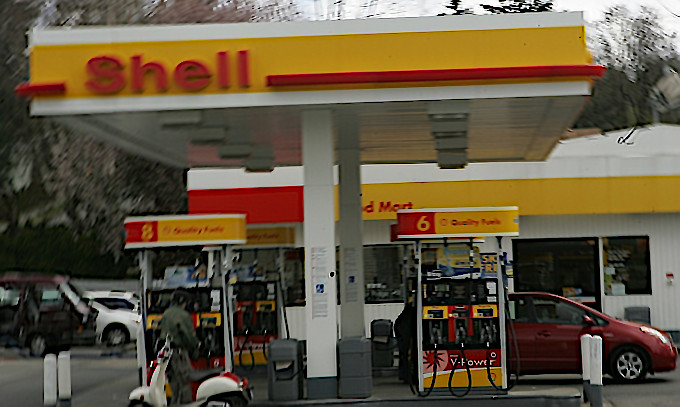}\\
& \Textcite{xu_cvpr2013} &Our result\\
\end{tabular}
\caption{Comparison on real-world
  example images taken from the literature with spatially-varying blur.}
\label{fig:bn.realworld_spatially}
\end{figure}

In Figs.\nobreakspace \ref {fig:bn.realworld_invariant} and\nobreakspace  \ref {fig:bn.realworld_spatially} we show results of our method on real-world images.
Figure\nobreakspace \ref {fig:bn.realworld_invariant} shows examples for invariant blur, while Fig.\nobreakspace \ref {fig:bn.realworld_spatially} depicts images with spatially-varying camera
shake. In both examples, our approach is able to recover images comparable
in quality with the state of the art.

\section{Discussion}

\subsection{Learned filters}

\begin{figure}[p]
\tightcolsep%
\small
\centering
\includegraphics[width=0.95\singlepw]{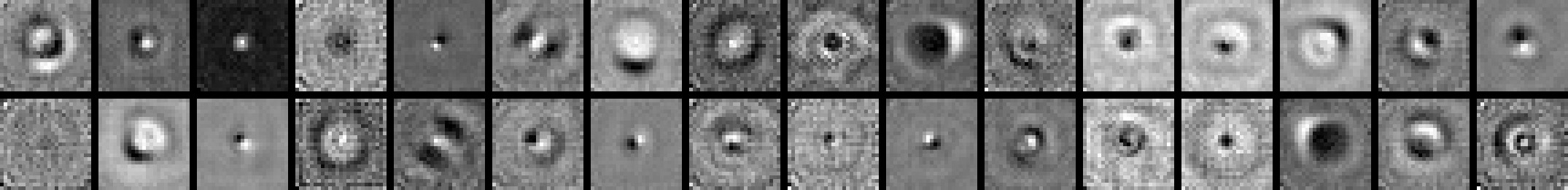}\\
First iteration\\[1.5ex]
\includegraphics[width=0.95\singlepw]{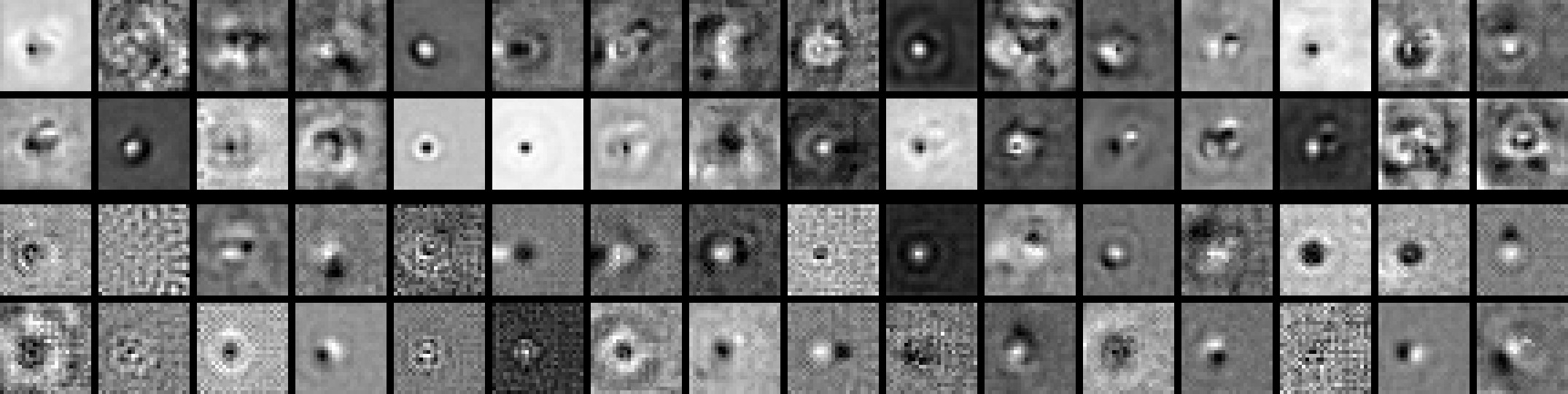}\\
Second iteration\\[1.5ex]
\includegraphics[width=0.95\singlepw]{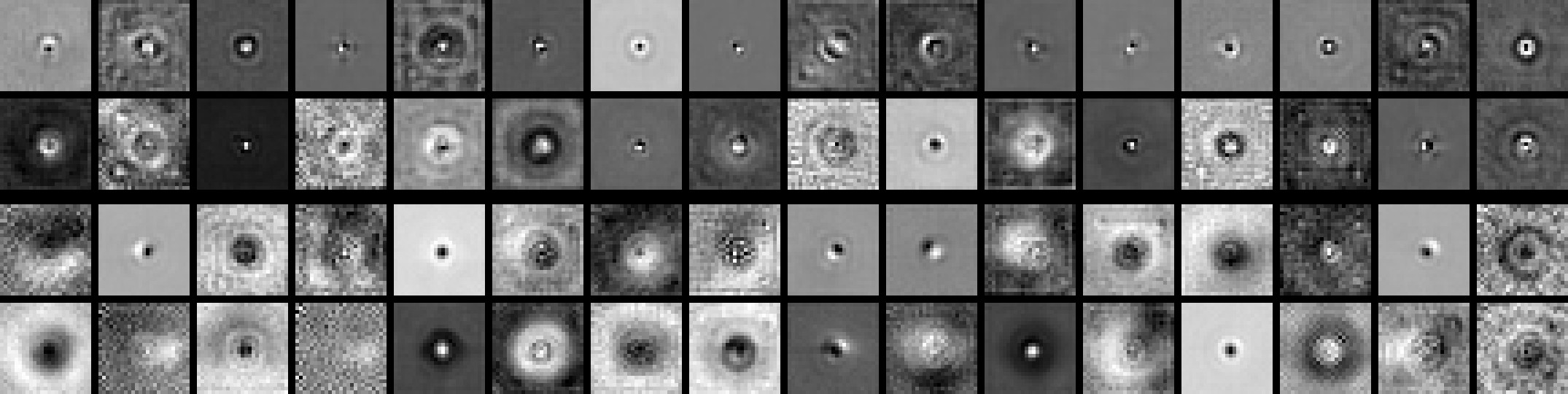}\\
Third iteration
\caption[Learned filters of the convolution layer for each of the
  three iterations within a single scale of a trained NN.]{Learned filters of the convolution layer for each of the
  three iterations within a single scale of a trained NN for
  kernel size 17$\times$17 pixels. See text for details.}
\label{fig:filters}
\end{figure}

\begin{figure}[t]
  \setlength{\fboxsep}{0pt}%
  \setlength{\fboxrule}{0.5pt}%
  \centering
  \tightcolsep%
  \small
  \begin{tabular}{ccccc}
  \fbox{\includegraphics[width=\quintpw]{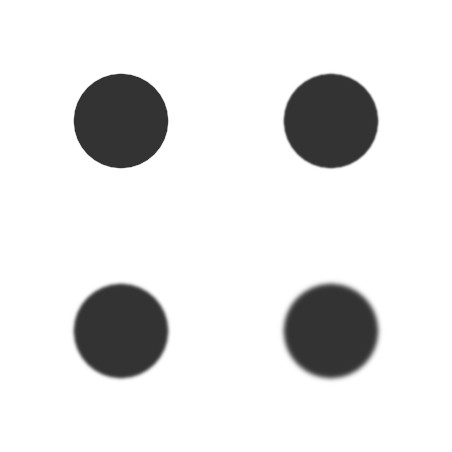}}&
  \fbox{\includegraphics[width=\quintpw]{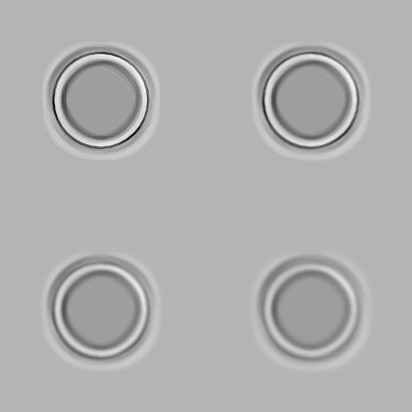}}&
  \fbox{\includegraphics[width=\quintpw]{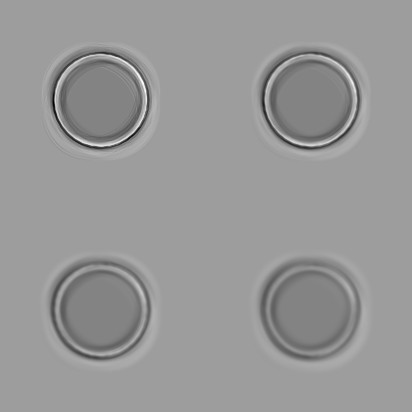}}&
  \fbox{\includegraphics[width=\quintpw]{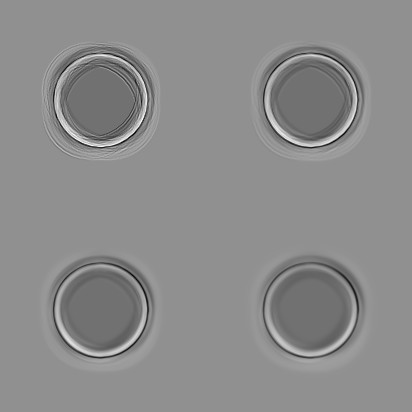}}&
  \fbox{\includegraphics[width=\quintpw]{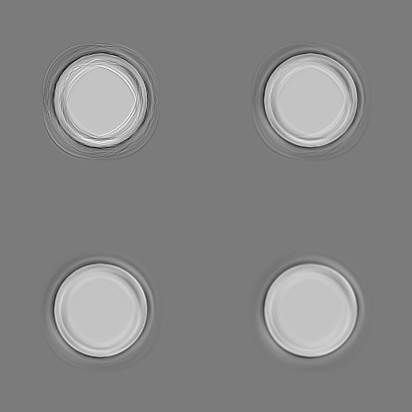}}\\
  \fbox{\includegraphics[width=\quintpw]{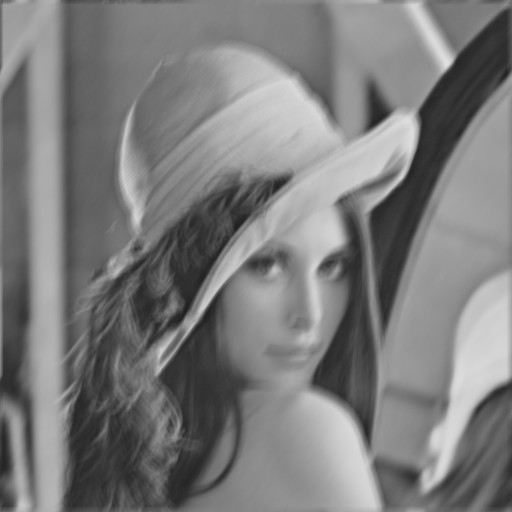}}&
  \fbox{\includegraphics[width=\quintpw]{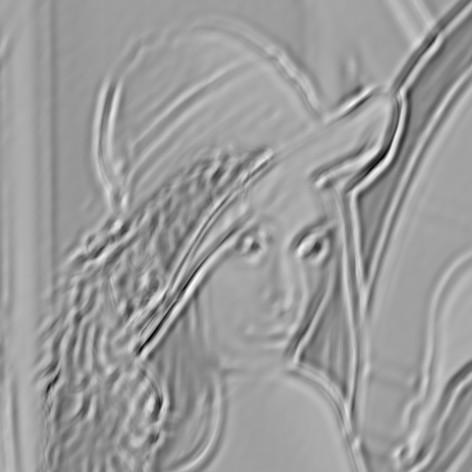}}&
  \fbox{\includegraphics[width=\quintpw]{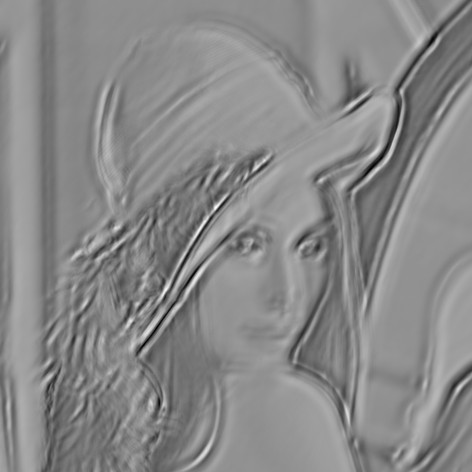}}&
  \fbox{\includegraphics[width=\quintpw]{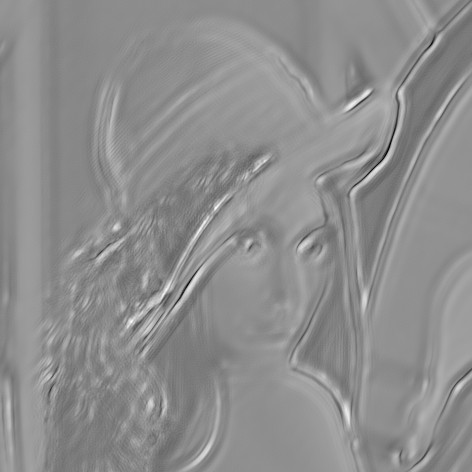}}&
  \fbox{\includegraphics[width=\quintpw]{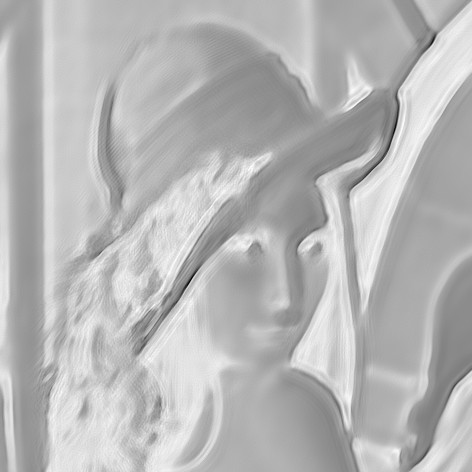}}\\
  Blurry image & Learned blurry   & Learned blurry & Learned latent &
  Learned latent \\
  & feature image $\tilde{\vv{y}}_1$  & feature image $\tilde{\vv{y}}_2$ & feature image $\tilde{\vv{x}}_1$  &  feature image $\tilde{\vv{x}}_2$
  \end{tabular}
  \caption[Visualization of the effect of the first stage of a
    network with two predicted output images on toy example with
    disks blurred with Gaussians of varying size and motion
    blurred Lena image.]{Visualization of the effect of the first stage of a
    network with two predicted output images on toy example with
    disks blurred with Gaussians of varying size (top row) and motion
    blurred Lena image (bottom row). While for the circles in $\tilde{\vv{y}_i}$ the
    different sizes of the Gaussian blurs are clearly visible, the NN replaces
    them in $\tilde{\vv{x}_i}$ with shapes of comparable sharpness.}
\label{fig:effect}
\end{figure}

The task of the \emph{Feature Extraction Module} is to emphasize and
enhance those image features that contain information about the
unknown blur kernel. Figure\nobreakspace \ref {fig:filters} shows the learned
filters of the convolution layer for each of the three stages within a
single scale of a trained NN for kernel size 17$\times$17
pixels. While the first stage takes a single (possibly down-sampled)
version of a blurry image as input, the subsequent stages take both
the restored latent image (obtained by non-blind deconvolution with
the current estimate of the kernel) and the blurry image as input. The
outputs of each stage are nonlinearly filtered versions of the input
images. In Fig.\nobreakspace \ref {fig:effect} we visualize the effect of the first
stage of a NN with two predicted output images on both the
Lena image and a toy example image consisting of four disks blurred with
Gaussians of varying size. Note that our feature extraction module outputs
nonlinearly filtered images for both the blurry and the latent sharp image,
both of which serve as input to the subsequent quotient layer, which in turn computes
an estimate of the blur kernel. This is in contrast to other existing
approaches \autocite{cho2009fast,Xu2010}, which apply a \emph{nonlinear} filter
to the current estimate of the latent image, but use only a \emph{linearly}
filtered version of the blurry input image for kernel estimation.

Once these feature images have passed the
subsequent tanh and recombination layer, they serve as input to
the \emph{quotient layer}, which computes an estimate of the unknown blur
kernel. Figure\nobreakspace \ref {fig:bn.intermediary} shows the intermediary results directly after
the convolution layer and how they progressively change after passing through
tanh and linear recombination layer two times, which seem to emphasize strong
edges.

\begin{figure}[p]
\tightcolsep%
\small
\centering
\includegraphics[width=0.95\singlepw]{input_filters/standard_broad}\\
NN trained on all images\\[1.5ex]
\includegraphics[width=0.95\singlepw]{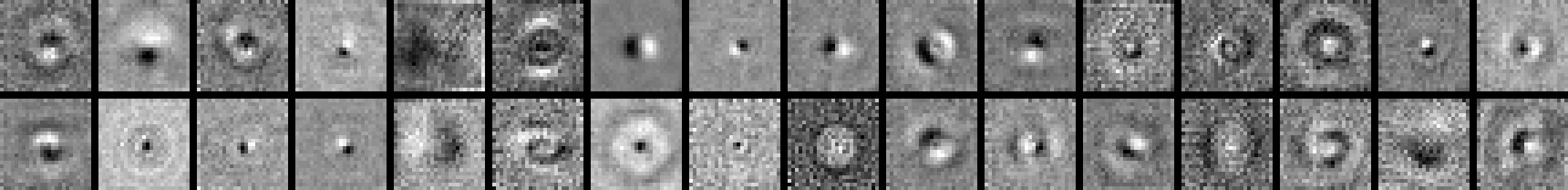}\\
NN trained on valley images\\[1.5ex]
\includegraphics[width=0.95\singlepw]{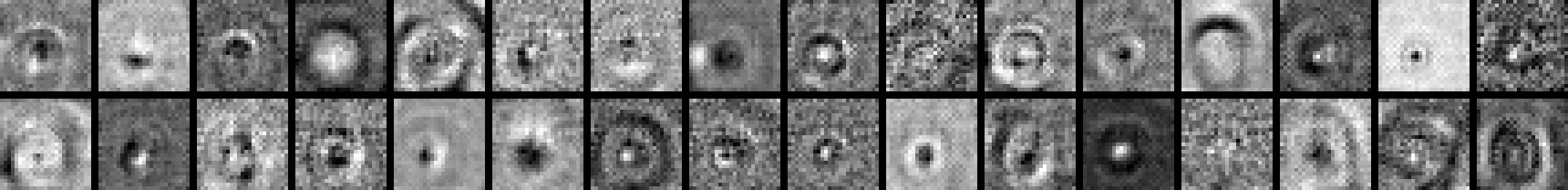}\\
NN trained on blackboard images
\caption[Comparison of learned filters.]{Learned filters of the convolution layer for three different types of images.}
\label{fig:bn.specific_filters}
\end{figure}

While the learned filters of the first stage are reminiscent of
gradient filters of various extent and orientations including Gabor and
Laplace-like filters, the filters of the subsequent stages are much
more intricate and more difficult to interpret.

As discussed in Section\nobreakspace \ref {sec:experiments}, the learned filters
depend on the image set that the NN was trained with,
\ie, the \emph{feature extraction module} learns image content
specific features that are informative about the unknown blur
kernel. In Fig.\nobreakspace \ref {fig:bn.specific_filters}, we show the learned filters of the
experiments in Section\nobreakspace \ref {sec:expert.training} for the \emph{valley} and
\emph{blackboard} image category of the ImageNet dataset; they indeed differ from
the ones trained on all images. For example, most of the filters learned for
valley images are mirror or rotational symmetric, unlike many filters
of the generic NN.

\subsection{Dependence on the size of the observed image}

As noted by \textcite{hu_eccv2012}, blind deblurring methods are most successful
in predicting the kernel in regions of an image that exhibit strong salient
edges. Other regions are less informative about the kernel, and have been shown
to even hurt kernel estimation when included in the input to the blind
deconvolution algorithm. Ideally, an estimation procedure should weight its
input according to its information content. In the worst case, a larger input
would not improve the results, but would not cause a deterioration either.

\begin{figure}
 \centering
 \small
 \renewcommand\tabcolsep{0.011\textwidth}%
 \begin{tabular}{ccccccc}
 $\vcenter{\hbox{\includegraphics[width=0.2\textwidth*\real{2}/\real{7}]{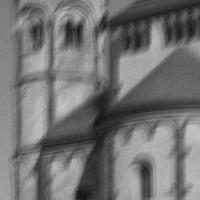}}}$&
 $\vcenter{\hbox{\includegraphics[width=0.2\textwidth*\real{3}/\real{7}]{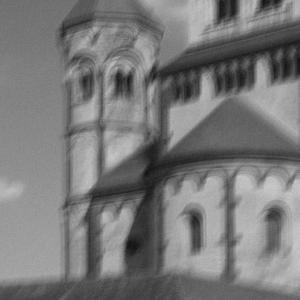}}}$&
 $\vcenter{\hbox{\includegraphics[width=0.2\textwidth*\real{4}/\real{7}]{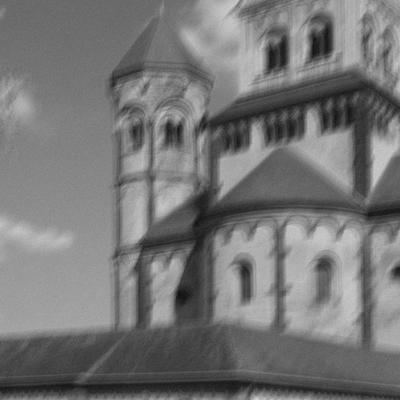}}}$&
 $\vcenter{\hbox{\includegraphics[width=0.2\textwidth*\real{5}/\real{7}]{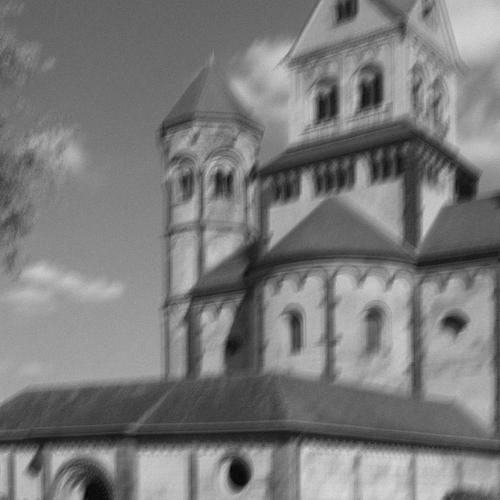}}}$&
 $\vcenter{\hbox{\includegraphics[width=0.2\textwidth*\real{6}/\real{7}]{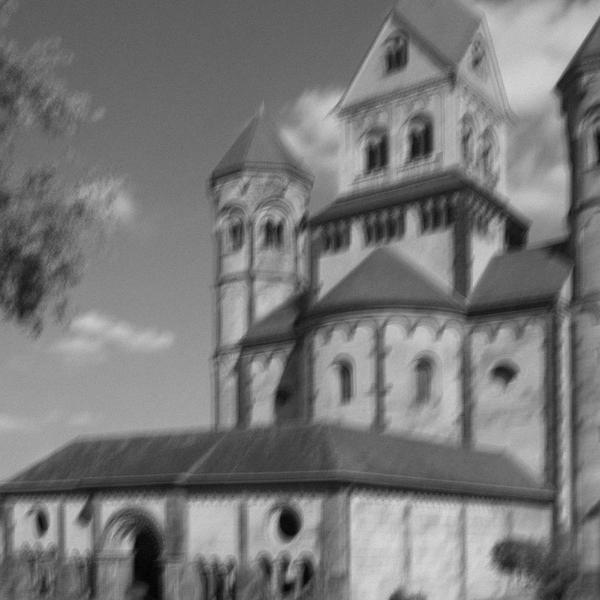}}}$&
 $\vcenter{\hbox{\includegraphics[width=0.2\textwidth*\real{7}/\real{7}]{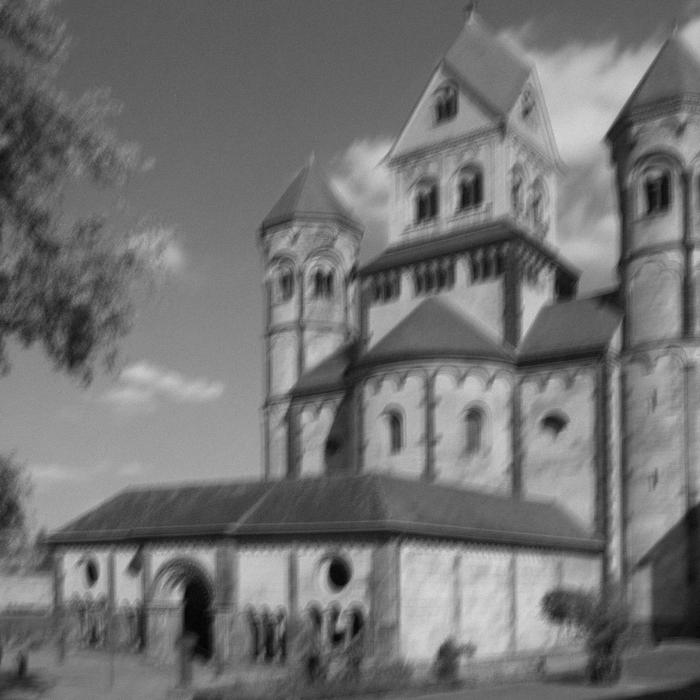}}}$
 \\
 \includegraphics[angle=180,width=0.1\textwidth]{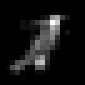}&
 \includegraphics[angle=180,width=0.1\textwidth]{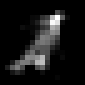}&
 \includegraphics[angle=180,width=0.1\textwidth]{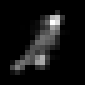}&
 \includegraphics[angle=180,width=0.1\textwidth]{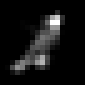}&
 \includegraphics[angle=180,width=0.1\textwidth]{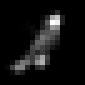}&
 \includegraphics[angle=180,width=0.1\textwidth]{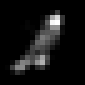}
 \\
 \includegraphics[angle=180,width=0.1\textwidth]{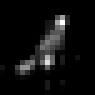}&
 \includegraphics[angle=180,width=0.1\textwidth]{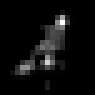}&
 \includegraphics[angle=180,width=0.1\textwidth]{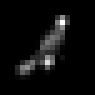}&
 \includegraphics[angle=180,width=0.1\textwidth]{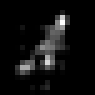}&
 \includegraphics[angle=180,width=0.1\textwidth]{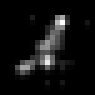}&
 \includegraphics[angle=180,width=0.1\textwidth]{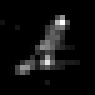}
 \end{tabular}
 \caption[Results for kernel estimation for different sizes of the observed
 image.]{Results for kernel estimation for different sizes of the observed image.
 \emph{Top row:} Differently sized inputs to the blind deblurring algorithm.
 \emph{Middle row:} Estimated kernels of our method. Larger inputs lead to better
 results. \emph{Bottom Row:} Estimated kernels of \autocite{xu_cvpr2013}.
 No clear trend is visible.}
\label{fig:bn.inputsize_example}
\end{figure}

\begin{figure}
  \centering
  \small
  \includegraphics{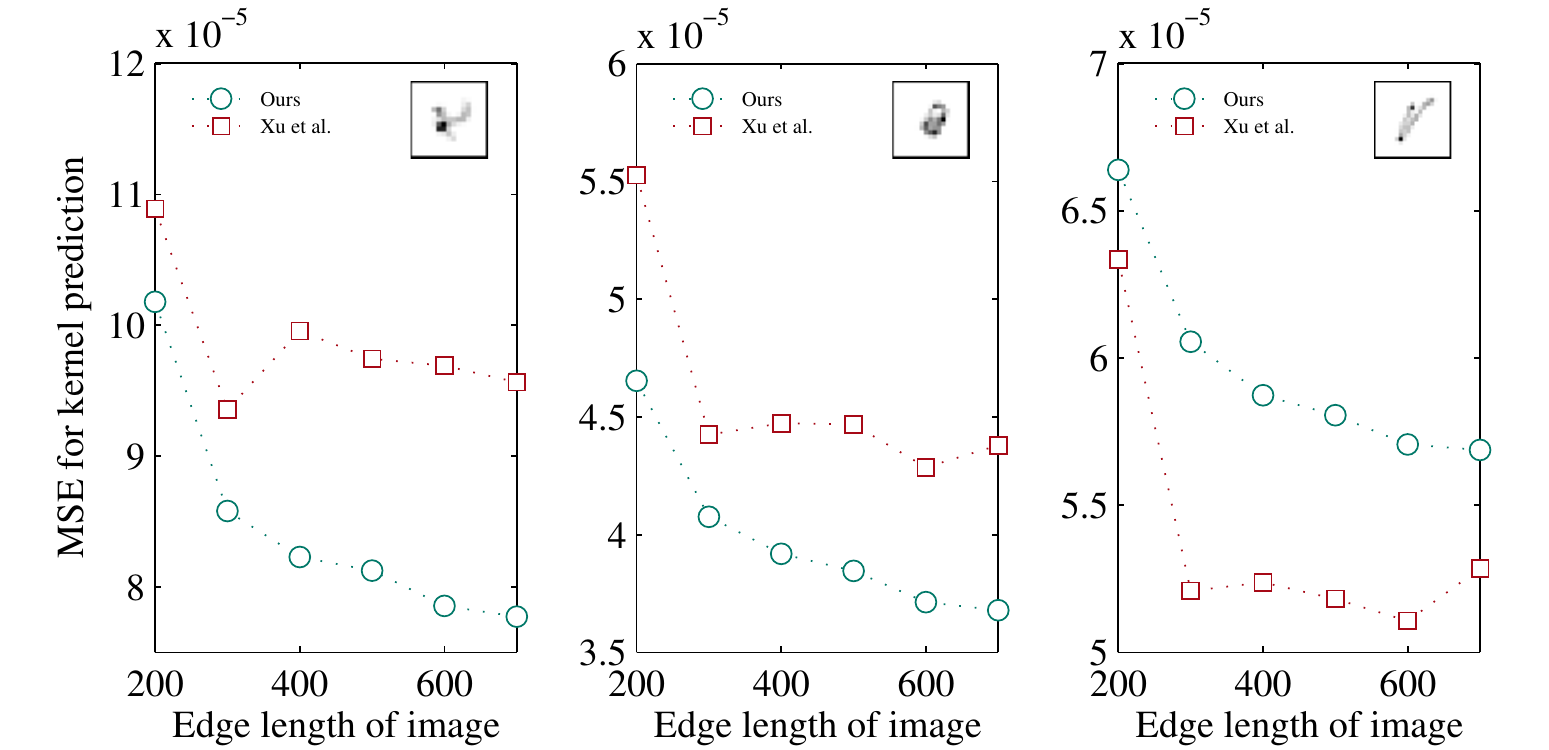}
 \caption[Dependence of the estimated kernel on the size of the observed image.]{Dependence of the estimated kernel on the size of the observed image. We show
 the MSE of predictions of three kernels for different sized inputs (\cf
 Fig.\nobreakspace \ref {fig:bn.inputsize_example}), averaged over the 52 largest images from
 \autocite{sun2013}. The results of the NN decrease monotonously.}
\label{fig:bn.inputsize}
\end{figure}

We study the behavior of our method with respect to the size of the observed
image. It is possible that the NN learned to ignore image content
detrimental to the kernel estimation. In Fig.\nobreakspace \ref {fig:bn.inputsize_example} the
predicted kernels for different sized crops of a blurry image are shown. Indeed,
this example suggests that for our learned algorithm the kernel converges with
input images increasing in size, while a non-learned state-of-the-art algorithm
\autocite{xu_cvpr2013} exhibits no such trend. The more thorough analysis in
Fig.\nobreakspace \ref {fig:bn.inputsize} confirms this behavior: when evaluating the MSE
for three different kernels, averaged over the 52 largest images from
\autocite{sun2013}, it monotonously decreases for larger crops of the blurry
image. We also see that the NN outperforms the competing method for
the two small blur kernels.

\subsection{Limitations}
\label{sec:failure}

\begin{figure}
\centering
\renewcommand\tabcolsep{1em}%
\small
\begin{tabular}{ccc}
\raisebox{0.02\quadpw}{\includegraphics[width=0.8\triplepw]{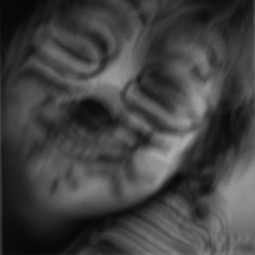}}\hspace*{0.1\quadpw}
&
\begin{tikzpicture}[every node/.style={anchor=south east,inner sep=0pt}]
\node (fig1) at (0,0){\includegraphics[width=0.8\triplepw]{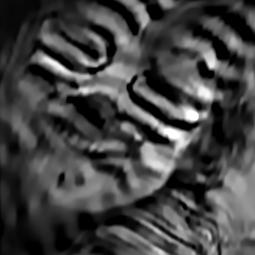}};%
\node (fig2) at (0.1\quadpw,-0.02\quadpw){\includegraphics[width=0.3\quadpw]{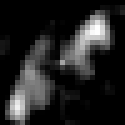}};%
\end{tikzpicture}
&
\begin{tikzpicture}[every node/.style={anchor=south east,inner sep=0pt}]
\node (fig1) at (0,0){\includegraphics[width=0.8\triplepw]{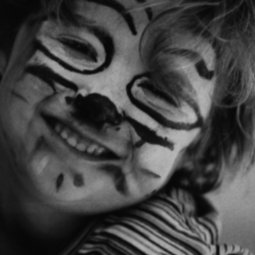}};%
\node (fig2) at (0.1\quadpw,-0.02\quadpw){\includegraphics[width=0.3\quadpw]{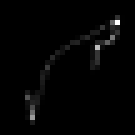}};%
\end{tikzpicture}
\\
Blurry image & Failure case of our approach & Ground truth
\end{tabular}
\caption[Failure case of our approach.]{For larger blur kernels our approach falls short in
  yielding acceptable deblurring results. Shown is an example from~\autocite{levin2009understanding} with a kernel of size 27$\times$27.} 
\label{fig:failure}
\end{figure}

A limitation of our current approach is the performance drop in the
case of larger blur kernels. Figure\nobreakspace \ref {fig:failure} shows an example
failure case from the benchmark dataset of~\autocite{levin2009understanding}. We believe
the reason for this is the suboptimal architecture of our
multi-scale approach at higher resolution scales. While a multi-scale
approach exhibits better performance compared to a single scale
network, the observed performance drop for larger blurs suggests that
using the same architecture at all scales is not
optimal.

\section{Conclusion}

We have shown that it is possible to automatically learn blind deconvolution by
reformulating the task as a single large nonlinear regression problem, mapping
between blurry input and predicted kernel. The key idea is to incorporate
the properties of the generative forward model into our algorithm, namely that
the image is convolved with the same blur kernel everywhere. While features are
extracted locally in the image, the \emph{kernel estimation module} combines
them globally. Next, the \emph{image estimation module} propagates the information to the
whole image, reducing the difficulty of the problem for the following iteration.

Our approach 
can adapt to different settings (\eg, blurry images with strong noise,
or specific image classes), and it could be further extended to combine deblurring with
other steps of the imaging pipeline, including over-saturation,
Bayer filtering, HDR, or super-resolution.

The blur class also invites future research: instead of artificially sampling from a stochastic process,
one could use recorded spatially-varying camera shakes, or a different source of
unsharpness, like lens aberrations or atmospheric turbulences in
astrophotography. Additionally, the insights gained from the trained system
could be beneficial to existing hand-crafted methods. This includes using
higher-order gradient representations and extended gradient filters.

Scalability of our method to large kernel sizes is still an issue, and this
may benefit from future improvements in neural net architecture and training.

\begin{appendices}
\pdfoutput=1

\section{Quotient Layer}
\label{app:math_details}
As introduced in Eq.\nobreakspace \textup {(\ref {eq:bn.quotient_layer})}, the quotient layer performs the operation
\begin{equation}
\tilde{\vv{k}} = \F^\H\frac{\sum_i \conj{\F\tilde{\vv{x}_i}}\odot \F\tilde{\vv{y}_i}}
  {\sum_i|\F\tilde{\vv{x}_i}|^2 + \beta_k}
\end{equation}
to estimate the kernel $\tilde{\vv{k}}$ from images $\tilde{\vv{x}_i}$ and their blurry
counterparts $\tilde{\vv{y}_i}$, both predicted by the previous layers of the
NN\@.

The quotient layer also includes a learned regularization parameter
$\beta_k$. To train the
NN, we need the gradients of the output with respect to
its parameters (in this case $\tilde{\vv{x}_i}$, $\tilde{\vv{y}_i}$ and
$\beta_k$) in every layer. From these we obtain the gradient steps $\Delta
\tilde{\vv{x}_j}$, $\Delta \tilde{\vv{y}_k}$ and $\Delta \beta_k$ in terms of
$\Delta \tilde{\vv{k}}$, where $\Delta \tilde{\vv{k}}$ is determined by
the loss for the current training
example, back-propagated through the layers subsequent to the quotient layer.

\subsection{Derivative with respect to sharp images}

To obtain $\Delta \tilde{\vv{x}_j}$, we first determine the differential form in
numerator layout, which means for vectors $\vv{u}$, $\vv{v}$ and a matrix $M$ that
\begin{equation}
d\vv{u}=\begin{pmatrix}
du_1\\
du_2\\
\vdots
\end{pmatrix}
,\quad
\frac{d\vv{u}}{d\vv{v}}=\begin{pmatrix}
\frac{du_1}{dv_1} & \frac{du_1}{dv_2} & \cdots\\
\frac{du_2}{dv_1} & \frac{du_2}{dv_2} & \cdots\\
\vdots  & \vdots  & \ddots
\end{pmatrix}
\quad \text{and} \quad
dM=\begin{pmatrix}
dm_{11} & dm_{12} & \cdots\\
dm_{21} & dm_{22} & \cdots\\
\vdots  & \vdots  & \ddots
\end{pmatrix}.
\end{equation}
Therefore, assuming that only $\tilde{\vv{x}_j}$ is variable, with the rules of
matrix calculus \autocite{petersen2012matrix} we arrive at
\begin{align}
d\tilde{\vv{k}} &=
  \F^\H \frac{\conj{\F d\tilde{\vv{x}_j}}\odot \F\tilde{\vv{y}_j}}
  {\sum_i|\F\tilde{\vv{x}_i}|^2 + \beta_k}-\F^\H\frac{(\sum_i \conj{\F\tilde{\vv{x}_i}}\odot \F\tilde{\vv{y}_i})\odot d(\F\tilde{\vv{x}_j}\odot\overline{\F\tilde{\vv{x}_j}})}
  {{(\sum_i|\F\tilde{\vv{x}_i}|^2 + \beta_k)}^2}=\notag\\
  &=\underbrace{\F^\H\Diag\left(\frac{\F\tilde{\vv{y}_j}}{\sum_i|\F\tilde{\vv{x}_i}|^2 + \beta_k}\right)\conj{\F}}_A d\tilde{\vv{x}_j}\notag\\
    &\phantom{=}-\underbrace{\F^\H\Diag\left(\frac{(\sum_i \conj{\F\tilde{\vv{x}_i}}\odot \F\tilde{\vv{y}_i})\odot \overline{\F\tilde{\vv{x}_j}}}{{(\sum_i|\F\tilde{\vv{x}_i}|^2 + \beta_k)}^2}\right) \F}_B d\tilde{\vv{x}_j}-\underbrace{\F^\H\Diag\left(\frac{(\sum_i \conj{\F\tilde{\vv{x}_i}}\odot \F\tilde{\vv{y}_i})\odot \F\tilde{\vv{x}_j}}{{(\sum_i|\F\tilde{\vv{x}_i}|^2 + \beta_k)}^2}\right)\overline{\F}}_C d\tilde{\vv{x}_j},
\end{align}
with the Hadamard product $\vv{a}\odot \vv{b} = \Diag(\vv{a})\vv{b}$ and the operation
``$\Diag$'' that creates a diagonal matrix from a vector. Additionally,
$\tilde{\vv{x}_i}$ is real, and therefore $\conj{d\tilde{\vv{x}_j}}=d\tilde{\vv{x}_j}$.

With the differential $d\tilde{\vv{k}}=M d\tilde{\vv{x}_j}$ (where $M=A+B+C$) in numerator layout, we note that the derivative $\frac{d\tilde{\vv{k}}}{d\tilde{\vv{x}_j}}=M$ and the gradient step
$\Delta \tilde{\vv{x}_j}=M^\T \Delta \tilde{\vv{k}}$.
Additionally, we use that $B^\T \Delta \tilde{\vv{k}}$ is real since it is the
inverse Fourier transform of a point-wise product of vectors all with Hermitian
symmetry (they are themselves purely real vectors that have been Fourier
transformed). Hermitian symmetry for a vector $\vv{v}$ of length $n$ means
$\vv{v}_{n-i+1}=\conj{\vv{v}_i}$ for all its components $i$. Therefore,
\begin{align}
\Delta \tilde{\vv{x}_j} &= A^\T \Delta \tilde{\vv{k}} - B^\T \Delta \tilde{\vv{k}} - C^\T \Delta \tilde{\vv{k}} = A^\T \Delta \tilde{\vv{k}} - \overline{B^\T \Delta \tilde{\vv{k}}} - C^\T \Delta \tilde{\vv{k}}=\notag\\
  &=\F^\H \frac{\overline{\F\Delta \tilde{\vv{k}}}\odot\F\tilde{\vv{y}_j}}{\sum_i|\F\tilde{\vv{x}_i}|^2 + \beta_k}\notag\\
  &\phantom{=}-\F^\H \frac{\overline{\sum_i \conj{\F\tilde{\vv{x}_i}}\odot \F\tilde{\vv{y}_i}}\odot\F\tilde{\vv{x}_j}\odot \F \Delta \tilde{\vv{k}}}{{(\sum_i|\F\tilde{\vv{x}_i}|^2 + \beta_k)}^2}
    -\F^\H \frac{(\sum_i \conj{\F\tilde{\vv{x}_i}}\odot \F\tilde{\vv{y}_i})\odot\F\tilde{\vv{x}_j}\odot \overline{\F \Delta \tilde{\vv{k}}}}{{(\sum_i|\F\tilde{\vv{x}_i}|^2 + \beta_k)}^2}
    \notag\\
  &=\F^\H \left(
     \frac{\overline{\F\Delta \tilde{\vv{k}}}\odot\F\tilde{\vv{y}_j}}{\sum_i|\F\tilde{\vv{x}_i}|^2 + \beta_k} -
    \frac{2\Re\left(\overline{\sum_i \conj{\F\tilde{\vv{x}_i}}\odot \F\tilde{\vv{y}_i}}\odot\F\Delta \tilde{\vv{k}}\right)\odot\F\tilde{\vv{x}_j}}
    {{(\sum_i |\F\tilde{\vv{x}_i}|^2 + \beta_k)}^2}\right),
\end{align}
since the transpose has no effect on a diagonal matrix. The real part of a vector $\vv{v}$ is denoted as $\Re(\vv{v}) = \frac{1}{2}(\vv{v} + \conj{\vv{v}})$.

\subsection{Derivative with respect to blurry images}

The derivation for the gradient $\Delta \tilde{\vv{y}_j}$ is similar.
We again start with the differential form
\begin{equation}
d\tilde{\vv{k}} =
  \F^\H{}
  \frac{\conj{\F\tilde{\vv{x}_j}}\odot \F d\tilde{\vv{y}_j}}{\sum_i|\F\tilde{\vv{x}_i}|^2 + \beta_k}
  =
  \underbrace{\F^\H \Diag \left(\frac{\conj{\F\tilde{\vv{x}_j}}}
  {\sum_i|\F\tilde{\vv{x}_i}|^2 + \beta_k}\right)\F}_A d\tilde{\vv{y}_j},
\end{equation}
this time assuming $d\tilde{\vv{x}_j}$ and $d\beta_k$ to be zero.
Next, we use again that a real vector is unchanged by conjugation, and we
obtain
\begin{equation}
\Delta \tilde{\vv{y}_j} = A^\T \Delta\tilde{\vv{k}} = \overline{A^\T \Delta\tilde{\vv{k}}}
  = \F^\H \overline{\Diag \left(\frac{\conj{\F\tilde{\vv{x}_j}}}
  {\sum_i|\F\tilde{\vv{x}_i}|^2 + \beta_k}\right)}\F\Delta \tilde{\vv{k}}
  = \F^\H \left(\frac{{\F\tilde{\vv{x}_j}\odot \F\Delta \tilde{\vv{k}}}}
  {\sum_i|\F\tilde{\vv{x}_i}|^2 + \beta_k}\right).
\end{equation}

\subsection{Derivative with respect to the regularization parameter}

Following the previous procedure for the regularization parameter $\beta_k$ we arrive at
\begin{equation}
d\tilde{\vv{k}} = \F^\H{}
  \underbrace{\frac{\sum_i \conj{\F\tilde{\vv{x}_i}}\odot \F\tilde{\vv{y}_i}}
     {{(\sum_i|\F\tilde{\vv{x}_i}|^2 + \beta_k)}^2}}_A d\beta_k,
\end{equation}
assuming only $\beta_k$ to be variable. Then, multiplying $\Delta \tilde{\vv{k}}$ by
the transpose of $\frac{d\tilde{\vv{k}}}{d\tilde{\beta_k}}=A$ we get

\begin{equation}
\Delta \beta_k = A^\T \Delta \tilde{\vv{k}} =
  {\left(\F^\H{}
    \frac{\sum_i \conj{\F\tilde{\vv{x}_i}}\odot \F\tilde{\vv{y}_i}}
    {{(\sum_i|\F\tilde{\vv{x}_i}|^2 + \beta_k)}^2}
    \right)}^T \Delta \tilde{\vv{k}}.
\end{equation}

\end{appendices}
\printbibliography              
\end{document}